%% file: arxiv.tex
\documentclass[11pt]{article} 
\usepackage{latentgrm_preprint,times}

\input{math_commands.tex}

\usepackage{url}

\usepackage{graphicx}
\usepackage{booktabs}
\usepackage{float}
\usepackage{flafter} 
\usepackage{placeins}
\usepackage{xcolor}
\usepackage{colortbl}
\usepackage{listings}
\usepackage[ruled,linesnumbered]{algorithm2e}
\usepackage{hyperref}
\hypersetup{
    colorlinks=true,
    linkcolor=preprintblue,
    citecolor=preprintblue,
    urlcolor=preprintblue,
    pdftitle={Judging in Latent Space: Efficient Generative Reward Modeling via Semantics-Preserving Compression},
    pdfauthor={Mingqing Yuan, Xiaobo Liang, Junwei Yang, Ziwei Chen, Zeren Zhang, Hejin Wang, Yubin Wang, Juntao Li}
}

\definecolor{algorithmcomment}{RGB}{0,128,128}
\definecolor{semanticgreen}{RGB}{28,130,82}
\definecolor{fixedred}{RGB}{186,52,58}
\definecolor{globalblue}{RGB}{49,105,171}
\definecolor{rubricorange}{RGB}{211,125,32}
\definecolor{finalpurple}{RGB}{117,78,166}

\newcommand{\semanticbar}{\textcolor{semanticgreen}{\textbf{\textbar}}}
\newcommand{\fixedbar}{\textcolor{fixedred}{\textbf{\textbar}}}
\newcommand{\examplebar}{\textcolor{gray!85!black}{\textbf{\large\textbar}}\allowbreak}
\newcommand{\fixedexamplebar}{\textcolor{fixedred}{\textbf{\large\textbar}}\allowbreak}
\newcommand{\best}[1]{\cellcolor{black!8}\textbf{#1}}

\title{Judging in Latent Space: Efficient Generative Reward Modeling via Semantics-Preserving Compression}

\author{%
Mingqing Yuan\textsuperscript{1}, Xiaobo Liang\textsuperscript{1}, Junwei Yang\textsuperscript{2}, Ziwei Chen\textsuperscript{3},\\
Zeren Zhang\textsuperscript{4}, Hejin Wang\textsuperscript{5}, Yubin Wang\textsuperscript{6}, Juntao Li\textsuperscript{1}\thanks{Corresponding author.}\\[0.65em]
{\normalfont\fontsize{10}{13}\selectfont\textsuperscript{1} Soochow University \quad \textsuperscript{2} University of Cambridge}\\
{\normalfont\fontsize{10}{13}\selectfont\textsuperscript{3} Chalmers University of Technology \quad \textsuperscript{4} Peking University}\\
{\normalfont\fontsize{10}{13}\selectfont\textsuperscript{5} Tsinghua University \quad \textsuperscript{6} The Hong Kong University of Science and Technology}%
}
\date{}

\begin{document}

\maketitle

\begin{tcolorbox}[preprintpanel,before skip=0pt,after skip=0pt]
\begin{abstract}
Reward modeling often requires jointly representing and reasoning over multiple evaluation criteria, yet verbalizing this process token by token can incur substantial inference cost.
Recent work on latent reasoning suggests that continuous states may support this computation more compactly.
We introduce LatentGRM, a latent evaluation framework built on semantic chunking, compression, and reconstruction.
By using the structure of rubric-guided evaluations to guide compression, LatentGRM learns compact continuous trajectories that support autonomous pairwise judgments without generating textual assessments.
A separate interpreter reconstructs evaluation text from these trajectories, providing an offline view of the information retained under compression.
Under matched training data and backbones, LatentGRM achieves competitive aggregate preference accuracy relative to explicit Supervised Fine-Tuning (SFT) judges at both 4B and 8B scales. Across four benchmark domains, LatentGRM-8B compresses evaluation trajectories by $8.9$--$9.2\times$ and reduces total judge inference time by $6.1$--$7.0\times$ at vote@5. Controlled rubric interventions show that criterion-dependent preference information is carried through the latent sequence. Together, these results demonstrate that continuous latent evaluation can substantially reduce inference cost while preserving competitive judgment quality.

\end{abstract}

\vspace{0.5em}
{\color{preprintblue!30}\hrule height 0.5pt}
\vspace{0.4em}
  {\small\setlength{\parskip}{0pt}
  \raisebox{-0.22em}{\includegraphics[height=1.1em]{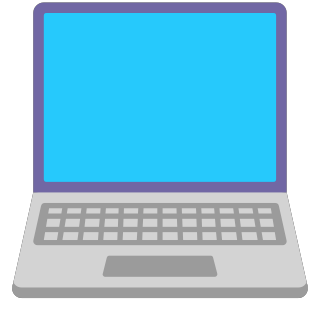}}\quad
  \textbf{Code}\quad\href{https://github.com/EADMO/LatentGRM}{https://github.com/EADMO/LatentGRM}\par
  \vspace{0.18em}
  \raisebox{-0.15em}{\includegraphics[height=1.0em]{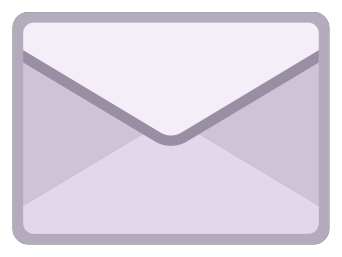}}\quad
  \textbf{Correspondence}\quad\href{mailto:mqyuan@stu.suda.edu.cn}{mqyuan@stu.suda.edu.cn}\quad\textbullet\quad\href{mailto:ljt@suda.edu.cn}{ljt@suda.edu.cn}\par}
\end{tcolorbox}

\begin{figure}[H]
    \centering
    \includegraphics[width=0.8\linewidth]{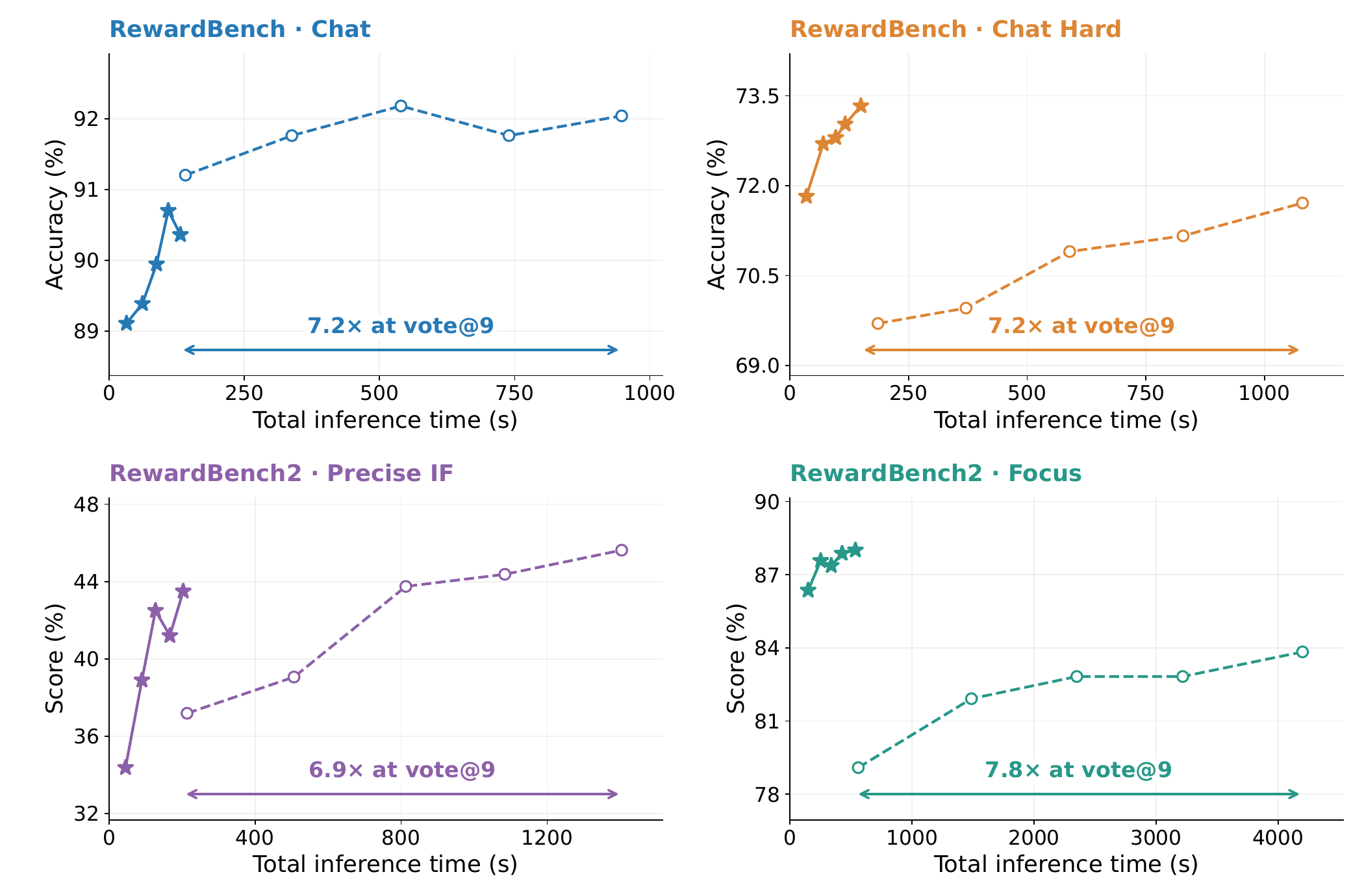}
    \caption{Accuracy--efficiency scaling of LatentGRM-8B (stars) and Rubric-RM-8B (circles) from vote@1 to vote@9. Points run left to right at 1, 3, 5, 7, and 9 votes.}
    \label{fig:accuracy_efficiency_scaling}
\end{figure}
\clearpage

\section{Introduction}
\label{sec:introduction}

Reward modeling often requires integrating evidence across multiple aspects of a response. A response may be factually correct yet violate an explicit constraint, while a detailed answer may fall short of a request for concision. These assessments must be combined into a preference that guides decisions in policy optimization and response selection~\citep{ouyang2022training,zhang2024generative}. Since reward models are used repeatedly in these settings, both judgment accuracy and inference efficiency are central to their practical value.

Among existing approaches, Generative Reward Models (GRMs) have become a dominant paradigm. Generative reward models formulate evaluation as language generation, with reasoning-based judges producing intermediate assessments before issuing a verdict~\citep{zhang2024generative,mahan2024generative,chen2025rmr1}. Recently, rubric-based reward modeling has gained increasing attention by organizing these assessments around \emph{rubrics}: explicit sets of task-specific evaluation criteria, such as correctness, instruction compliance, and clarity~\citep{liu2025openrubrics,yuan2026seeing}. While sampling and aggregating multiple evaluations can improve judgment quality, verbalizing judgments and supporting evidence for each criterion can lengthen evaluation trajectories and increase autoregressive decoding costs. This raises a central question: \emph{can a generative judge retain the information needed for multi-criterion evaluation while carrying out its intermediate computation more compactly?}

To address this, we turn to continuous latent reasoning, which offers a promising way to represent intermediate computation with fewer autoregressive steps~\citep{hao2024coconut,shen2025codi,deng2025latent}. Rubric-based reward modeling provides a particularly structured setting for studying such compact computation. Its criterion-level organization and linguistic boundaries can guide where to place compression boundaries, while the criteria themselves provide natural reference points for probing how latent judgments respond to changes in evaluation requirements and what assessment content remains recoverable.

We introduce \textbf{LatentGRM}, a framework for rubric-guided latent evaluation. Its organizing idea is to treat rubric criteria as \emph{intervenable and alignable criterion-level semantic units}. Building on Latent-SFT's vocabulary-space distillation framework~\citep{deng2025latent}, \textbf{Semantic Chunking} uses rubric-aligned blocks to allocate a fixed latent budget, then uses linguistic cues to refine boundaries within each block. The trained judge generates compact continuous trajectories and predicts pairwise preferences autonomously. A separate \textbf{Latent Trace Interpreter} (LTI) reconstructs evaluation text from saved trajectories outside the online judgment path. Aligning its output with teacher assessments by criterion provides a fine-grained measure of reconstruction fidelity on teacher-derived trajectories. Rubric structure therefore connects compression, intervention, and interpretation within one evaluation framework.

Against explicit Supervised Fine-Tuning (SFT) judges trained on matched data and 4B/8B backbones, LatentGRM achieves competitive aggregate preference accuracy across eight benchmark domains. On four domains, LatentGRM-8B compresses evaluation trajectories by $8.9$--$9.2\times$ and reduces total judge inference time by $6.1$--$7.0\times$ at vote@5. Controlled rubric edits and latent-source replacement show that criterion-dependent preference information is carried through autonomous latent trajectories. Separately, reconstruction experiments recover criterion-level states and justifications from teacher-derived trajectories. These findings support efficient latent judgment together with the recovery of useful evaluation content.

Our contributions are:
\begin{itemize}
    \item We introduce LatentGRM, to our knowledge the first framework to bring continuous latent reasoning to reward modeling, combining rubric-guided semantic compression, autonomous latent evaluation, and a separate interpreter for retrospective reconstruction.

    \item Under matched training data and backbones, we demonstrate competitive aggregate preference accuracy at both 4B and 8B scales and substantial reductions in trajectory length and total judge inference time.

    \item  We examine the evaluation information retained under compression through controlled rubric interventions and criterion-aligned reconstruction. These analyses show that latent trajectories carry criterion-dependent preference information and that the LTI can recover criterion-level supporting justifications from teacher-derived trajectories.
\end{itemize}

\section{Related Works}
\label{app:related-work}

\subsection{Latent Reasoning}

Continuous latent reasoning replaces explicit intermediate tokens with recurrent continuous states, reducing autoregressive decoding while retaining multi-step computation. Existing methods recurrently feed hidden states back into the model, distill textual reasoning chains into latent trajectories, or construct continuous thoughts in vocabulary space~\citep{hao2024coconut,shen2025codi,zhang2025softthinking,deng2025latent}. These methods have primarily been studied on mathematical and general reasoning tasks. We adapt vocabulary-space latent reasoning to reward modeling, where the latent trajectory must represent a multi-criterion comparison between candidate responses.

\subsection{Reward Modeling}
Generative reward models formulate evaluation as language generation and can produce critiques or reasoning traces before issuing a judgment~\citep{mahan2024generative,ankner2024cloud,chen2025rmr1,guo2025rewardreasoning}. In recent years, rubric-based reward modeling has rapidly gained attention as a way to decompose holistic judgments into explicit, task-specific criteria for evaluation, feedback, reward construction, and policy optimization~\citep{liu2025openrubrics,yuan2026seeing}. However, such rubric-based and reasoning-heavy judges introduce substantial inference costs, since they must generate and apply criteria explicitly at test time. Our work therefore focuses on the inference cost of reasoning-based generative judges: LatentGRM compresses the evaluations into continuous trajectories, with matched explicit SFT judges as the primary comparison. We further repurpose rubrics as anchors for semantic chunking and as an evaluation and analysis tool for latent reasoning.

\begin{figure}[t]
    \centering
    \includegraphics[width=\textwidth]{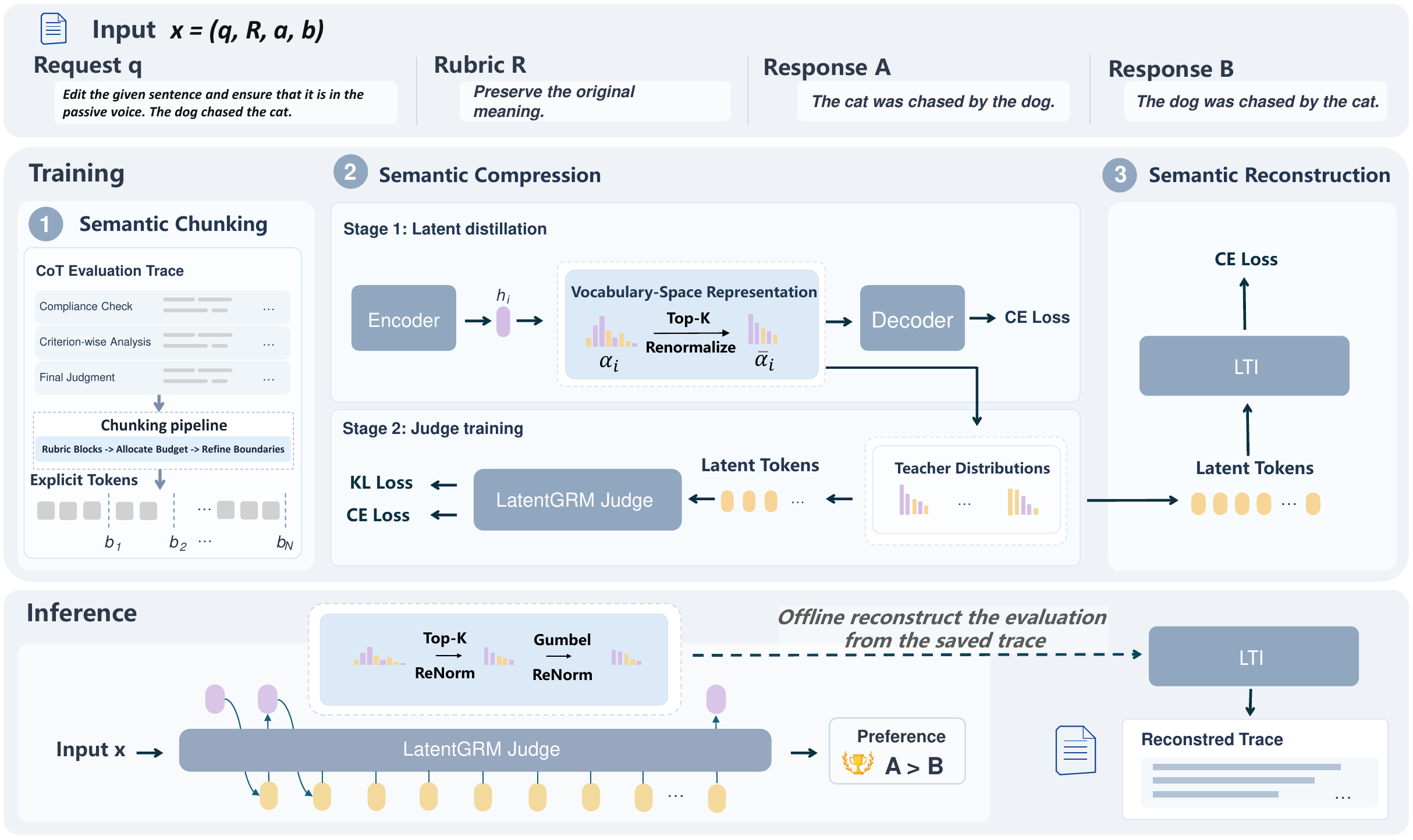}
    \caption{LatentGRM training and inference. Stage~1 encodes semantic boundaries into vocabulary-space targets and trains suffix reconstruction; Stage~2 distills autonomous latent judging. An offline interpreter reconstructs evaluations from saved trajectories.}
    \label{fig:latentgrm_overview}
\end{figure}

\section{Method}
\label{sec:method}

\subsection{Overview}

We study pairwise reward modeling with an evaluation input $x=(q,\mathcal{R},a,b)$, comprising a user request, a task-specific rubric, and two candidate responses. Training examples additionally provide an explicit evaluation trace $c=(c_1,\ldots,c_M)$ and a preference label $y\in\{\texttt{Response A},\texttt{Response B}\}$. An explicit reasoning judge produces:
\begin{equation}
    [x,\ \texttt{<think>},\ c_1,\ldots,c_M,\ \texttt{</think>},\ y].
    \label{eq:explicit-grm}
\end{equation}
LatentGRM replaces the textual evaluation with fewer continuous steps:
\begin{equation}
    [x,\ \texttt{<think>},\ z_1,\ldots,z_N,\ \texttt{</think>},\ y].
    \label{eq:latent-grm}
\end{equation}
We build on Latent-SFT's vocabulary-space representation and two-stage distillation framework~\citep{deng2025latent}, replacing fixed-rate segmentation with Semantic Chunking. Stage~1 distills explicit evaluation traces into latent trajectories. Stage~2 trains the judge to autoregressively generate these trajectories and predict the final preference from the evaluation input at inference. A separate interpreter reconstructs explicit evaluations from saved trajectories for offline analysis. Figure~\ref{fig:latentgrm_overview} summarizes the framework.

\subsection{Stage 1: Semantic Chunking for Latent Compression}
\label{sec:semantic-chunking}

Stage~1 follows an encoder and decoder objective. At each semantic boundary, the encoder compresses the evaluation input and teacher-trace prefix into latent states, while the decoder learns to recover the remaining evaluation and final preference from the latent prefix.

Let $c_{1:M}$ denote an evaluation trace containing $M$ tokens. Semantic Chunking divides it into $N$ spans using ordered boundaries $0=b_0<b_1<\cdots<b_N=M$. We set:
\begin{equation}
    N=\left\lceil\frac{M}{\rho}\right\rceil,
    \label{eq:fixed-latent-budget}
\end{equation}
where $\rho$ is the fixed compression rate. Rubric-guided evaluations naturally separate into compliance check, criterion-specific analysis, and final synthesis (Figure~\ref{fig:teacher-reference-case-14270}). We fix the latent budget of each rubric block with an exact-budget allocator, then refine its internal frontiers using sentence, clause, and dependency evidence. Semantic Chunking places compression boundaries to preserve semantic integrity as much as possible.

Figure~\ref{fig:segmentation-examples} illustrates how the two boundary choices treat the same rubric and sentence-level passages.
\input{figures/semantic_chunking_illustration.tex}

At boundary $b_i$, the encoder reads $x$ and the explicit prefix $c_{1:b_i}$ and produces a hidden state $h_i$. To make this state directly consumable by the decoder, we express it through the decoder's vocabulary interface. Let $W_{\mathrm{out}}$ be the decoder's frozen vocabulary output projection and $\tau$ the projection temperature. The resulting full-vocabulary distribution is $\alpha_i=\operatorname{softmax}(W_{\mathrm{out}}h_i/\tau)$. We define $S_i=\operatorname{TopK}(\alpha_i)$ as the indices of the $K$ largest entries of $\alpha_i$ and renormalize their probability mass, so $\bar\alpha_{i,v}=\alpha_{i,v}/\sum_{u\in S_i}\alpha_{i,u}$ for $v\in S_i$, and $\bar\alpha_{i,v}=0$ otherwise. Using the decoder's frozen input embedding table $E_{\mathrm{in}}$, we form the compressed state:
\begin{equation}
    z_i
    =
    \sum_{v\in S_i}
    \bar{\alpha}_{i,v}E_{\mathrm{in}}[v].
    \label{eq:vocabulary-latent}
\end{equation}
Thus $z_i$ is a continuous mixture of token embeddings and can be inserted directly into the decoder's input sequence. Given $[x,\texttt{<think>},z_{1:i}]$, the decoder predicts the remaining evaluation and final preference:
\begin{equation}
    \mathcal{L}_{\mathrm{rec}}
    =\frac{1}{N}\sum_{i=1}^{N}
    \operatorname{CE}\!\left(
    [c_{b_i+1:M},\texttt{</think>},y,\texttt{EOS}]
    \mid
    [x,\texttt{<think>},z_{1:i}]
    \right).
    \label{eq:latent-reconstruction}
\end{equation}
Here, $\operatorname{CE}$ is token-averaged autoregressive cross-entropy. We optimize the encoder, the decoder, and then both jointly. After training, the encoder exports $(S_i,\bar\alpha_i)_{i=1}^N$ as sparse targets for Stage~2.

\subsection{Stage 2: Autonomous Latent Evaluation}

Starting from the Stage~1 decoder, Stage~2 trains the judge to predict the exported distributions autoregressively and then output the final preference. Let $\mathbf g=(\mathbf g_1,\ldots,\mathbf g_N)$ collect independent Gumbel noise across the trajectory. For each target $\bar\alpha_i$, we add $\mathbf g_i$ to its log-probabilities and renormalize within $S_i$, obtaining $\widetilde\alpha_i(\mathbf g)$. Replacing $\bar\alpha_i$ with these perturbed weights in Eq.~\ref{eq:vocabulary-latent} gives the corresponding target embedding $\widetilde z_i(\mathbf g)$. During training, the prediction at step $i$ is conditioned on the preceding perturbed target embeddings $\widetilde z_{1:i-1}(\mathbf g)$. This stochastic perturbation discourages overfitting to deterministic teacher targets; at inference, the same mechanism produces diverse latent trajectories for voting. Let $x_{\mathrm{lat}}=[x,\texttt{<think>}]$ and let $q_\theta$ denote the judge's predicted full-vocabulary distribution. We optimize:
\begin{equation}
\begin{aligned}
\mathcal{L}_{\mathrm{auto}}=\mathbb{E}_{\mathbf g}\Bigg[
\frac{1}{N}\sum_{i=1}^{N}
\operatorname{KL}\!\left(
\widetilde{\alpha}_i(\mathbf g)\,\middle\|\,
q_\theta(\cdot\mid x_{\mathrm{lat}},\widetilde z_{<i}(\mathbf g))\right)\\
\qquad+\operatorname{CE}\!\left(
[\texttt{</think>},y,\texttt{EOS}]
\mid x_{\mathrm{lat}},\widetilde z_{1:N}(\mathbf g)\right)\Bigg].
\end{aligned}
\label{eq:autonomous-latent}
\end{equation}
The KL term distills each perturbed latent target, while cross-entropy predicts the end marker, preference label, and EOS from the complete trajectory. The expectation is estimated with one noise draw per example at each update.

At inference, the judge feeds back its own predictions. At step $i$, it produces $q_i=q_\theta(\cdot\mid x,\texttt{<think>},z_{<i})$. If the unperturbed argmax of $q_i$ is \texttt{</think>}, latent generation terminates and the preference is decoded greedily. Otherwise, the judge retains and renormalizes the top-$K$ probabilities of $q_i$, applies the same Gumbel perturbation, and substitutes the resulting support and weights into Eq.~\ref{eq:vocabulary-latent} to form $z_i$. We extend vLLM~\citep{kwon2023vllm} to support this soft-embedding feedback while retaining continuous batching, KV caching, and tensor parallelism.

\subsection{Retrospective Interpretation of Latent Trajectories}
\label{sec:latent-interpreter}
The Latent Trace Interpreter (LTI) is a separate readout that reconstructs an explicit evaluation from a completed latent trajectory. At position $i$, it receives saved top-$K$ vocabulary indices $S_i^{\mathrm{save}}$ and normalized weights $\bar q_i^{\mathrm{save}}$. Using its embedding table $E_{\mathrm{int}}$, LTI forms:
\begin{equation}
    z_i^{\mathrm{int}}=\sum_{v\in S_i^{\mathrm{save}}}\bar q_i^{\mathrm{save}}(v)E_{\mathrm{int}}[v],
    \label{eq:interpreter-input-mixture}
\end{equation}
the same vocabulary-mixture construction as Eq.~\ref{eq:vocabulary-latent}. Given the full sequence $z_{1:N}^{\mathrm{int}}$, LTI predicts the complete teacher evaluation with the token-averaged objective:
\begin{equation}
    \mathcal L_{\mathrm{int}}=
    \operatorname{CE}\!\left(
    [c_{1:M},\texttt{</think>}]
    \mid [x,\texttt{<think>},z_{1:N}^{\mathrm{int}}]
    \right).
    \label{eq:interpreter-objective}
\end{equation}
Unlike Stage~1's suffix reconstruction, LTI reconstructs the evaluation from its beginning; the preference label $y$ is excluded. Autonomous judging can save its predicted top-$K$ indices and weights for the same offline readout, without running LTI in the online decision path.

\section{Experiments}
\label{sec:experiments}

Our experiments address three questions: (RQ1) whether latent evaluation preserves preference accuracy across model scales and responds to controlled changes in rubric criteria; (RQ2) how compressing explicit evaluations into latent trajectories affects inference cost; and (RQ3) whether criterion-level assessments can be reconstructed from latent trajectories.

\begin{table*}[t]
    \centering
    \caption{\textbf{Pairwise preference evaluation (\%).} Avg.-8 is the mean of the eight benchmark scores. Shading mark the best score among the Rubric-RM and LatentGRM configurations.}
    \label{tab:main-results}
    \scriptsize
    \setlength{\tabcolsep}{2.0pt}
    \resizebox{0.98\linewidth}{!}{%
    \begin{tabular}{@{}lccccccccc@{}}
        \toprule
        \textbf{Model} & \multicolumn{2}{c}{\textbf{RewardBench}} & \multicolumn{2}{c}{\textbf{IF Evaluation}} & \textbf{RM-Bench} & \multicolumn{2}{c}{\textbf{RewardBench2}} & \textbf{HS3} & \textbf{Avg.-8} \\
        \cmidrule(lr){2-3}\cmidrule(lr){4-5}\cmidrule(lr){6-6}\cmidrule(lr){7-8}
        & \textbf{Chat} & \textbf{Hard} & \textbf{PPE} & \textbf{IFBench} & \textbf{Chat} & \textbf{Prec. IF} & \textbf{Focus} & & \\
        \midrule
        \multicolumn{10}{l}{\textit{White-box judge/reward LLMs (for reference only)}} \\
        JudgeLRM-7B & 93.2 & 56.4 & 47.9 & 47.5 & 55.0 & 29.7 & 39.9 & 60.5 & 53.7 \\
        RRM-7B & 88.0 & 70.6 & 52.5 & 55.2 & 59.8 & 29.4 & 63.9 & 63.7 & 60.4 \\
        RM-R1-7B (Qwen-2.5-Inst) & 94.4 & 72.3 & 56.1 & 57.8 & 65.8 & 35.6 & 80.0 & 69.3 & 66.4 \\
        RM-R1-7B (DeepSeek-Dist) & 86.2 & 67.0 & 52.5 & 56.8 & 62.1 & 23.8 & 56.9 & 63.3 & 58.6 \\
        \midrule
        \multicolumn{10}{l}{\textit{Rubric-RM}} \\
        Rubric-RM-4B, vote@1 & 89.1 & 67.8 & 60.2 & 61.7 & 61.6 & 32.2 & 77.0 & 66.8 & 64.5 \\
        Rubric-RM-4B, vote@5 & 91.1 & 69.3 & 61.9 & 62.5 & 61.7 & 40.6 & 80.5 & 68.7 & 67.0 \\
        Rubric-RM-8B, vote@1 & 89.8 & 69.8 & 61.2 & 59.9 & 61.3 & 31.9 & 77.9 & 68.1 & 65.0 \\
        Rubric-RM-8B, vote@5 & \best{91.8} & 70.5 & 62.9 & 61.5 & \best{62.1} & \best{45.0} & 81.1 & \best{70.0} & 68.1 \\
        \addlinespace[2pt]
        \multicolumn{10}{l}{\textit{LatentGRM (ours)}} \\
        LatentGRM-4B, vote@1 & 87.2 & 68.4 & 61.2 & 61.0 & 59.9 & 34.1 & 85.4 & 67.9 & 65.6 \\
        LatentGRM-4B, vote@5 & 88.8 & 70.3 & 61.9 & 63.2 & 62.0 & 41.6 & 86.5 & 68.6 & 67.8 \\
        LatentGRM-8B, vote@1 & 88.7 & 70.5 & 60.4 & 63.0 & 59.8 & 37.9 & 84.6 & 68.2 & 66.7 \\
        LatentGRM-8B, vote@5 & 89.9 & \best{73.0} & \best{63.1} & \best{63.3} & 61.9 & 43.4 & \best{87.5} & 69.6 & \best{69.0} \\
        \bottomrule
    \end{tabular}}
\end{table*}

\subsection{Experimental Setup}

\paragraph{Training data.}
We use 35,612 OpenRubrics~\citep{liu2025openrubrics} training records, each containing a request, two responses, a rubric, an evaluation trace, and a binary preference. OpenRubrics keeps generated rubrics whose judgments match the preference labels.

\paragraph{Evaluation data.}
We evaluate eight benchmark domains: RewardBench Chat and Chat Hard~\citep{lambert2024rewardbench}, PPE-IFEval~\citep{frick2025evaluate}, IFBench~\citep{pyatkin2025generalizing}, RM-Bench Chat~\citep{liu2025rmbench}, RewardBench~2 Precise IF and Focus~\citep{malik2025rewardbench2}, and HelpSteer3~\citep{wang2025helpsteer3}. \textbf{Avg.-8} is their unweighted mean.

\paragraph{Models and evaluation.}
Our primary baselines are explicit Rubric-RM trained by SFT on the same OpenRubrics records and Qwen3-4B or Qwen3-8B backbones~\citep{qwen3} as LatentGRM. This matched comparison isolates the effect of replacing explicit evaluation traces with compact latent trajectories. Both methods receive identical task-specific rubrics, generated once offline by OpenRubrics generator-8B. We report a single rollout (vote@1) and majority voting over five rollouts (vote@5). JudgeLRM, RRM, and two RM-R1 variants serve as additional reference points~\citep{chen2025judgelrm,guo2025rewardreasoning,chen2025rmr1}. We evaluate every model in both response orders and average the resulting scores.

\subsection{RQ1: Effectiveness and Rubric-Based Process Evaluation}
\label{sec:reward-effectiveness}

\subsubsection{Preference Accuracy}

At vote@5, LatentGRM scores 67.8 and 69.0 on Avg.-8 at 4B and 8B, compared with 67.0 and 68.1 for the matched explicit Rubric-RM (Table~\ref{tab:main-results}). Its largest advantage is on Focus (+6.0 and +6.4 points); across the other seven domains, Chat Hard and IFBench favor LatentGRM at both scales, RewardBench Chat favors Rubric-RM, and Precise IF changes direction between 4B and 8B. PPE, RM-Bench Chat, and HelpSteer3 differ by at most 0.4 points. Thus, compact latent evaluation retains broadly similar preference accuracy across domains, with a concentrated gain on Focus and Chat Hard. Voting raises LatentGRM's Avg.-8 from 65.6 to 67.8 at 4B (+2.2) and from 66.7 to 69.0 at 8B (+2.3), showing that multiple compact trajectories can improve the final judgment.

Appendix~\ref{app:direct-judge} provides a comparison with a direct judge that predicts only the final preference.

\subsubsection{Effect of Semantic Chunking}
\label{sec:chunking-ablation}

To assess the contribution of Semantic Chunking, we compare it with Fixed-Rate Chunking on Qwen3-8B. Both use the same training data, target compression rate $\rho=8$, latent representation, initialization, and training schedule.

\begin{table}[t]
    \centering
    \small
    \caption{\textbf{Semantic Chunking ablation on Qwen3-8B (\%).}}
    \begin{tabular}{lrrrr}
        \toprule
        & \multicolumn{2}{c}{\textbf{RewardBench}} & \multicolumn{2}{c}{\textbf{RewardBench2}} \\
        \cmidrule(lr){2-3}\cmidrule(lr){4-5}
        \textbf{Method} & \textbf{Chat} & \textbf{Hard} & \textbf{Prec. IF} & \textbf{Focus} \\
        \midrule
        \multicolumn{5}{l}{\textit{vote@1}} \\
        Fixed-Rate Chunking & \best{90.5} & 69.0 & 37.2 & 83.3 \\
        Semantic Chunking & 88.7 & \best{70.5} & \best{37.9} & \best{84.6} \\
        \midrule
        \multicolumn{5}{l}{\textit{vote@5}} \\
        Fixed-Rate Chunking & \best{90.6} & 70.8 & 39.5 & 85.8 \\
        Semantic Chunking & 89.9 & \best{73.0} & \best{43.4} & \best{87.5} \\
        \bottomrule
    \end{tabular}
    \label{tab:chunking-results}
\end{table}

Semantic Chunking improves Chat Hard, Precise IF, and Focus at both voting budgets (Table~\ref{tab:chunking-results}). At vote@5, the gains over Fixed-Rate Chunking are 2.2, 3.9, and 1.7 points, respectively, while Chat decreases by 0.7 points; the four-domain mean rises by 1.8 points. With the latent budget and training conditions held fixed, this pattern indicates that placing boundaries at semantic units improves compressed judgments most clearly on instruction-sensitive tasks.

\subsubsection{Rubric Interventions and Latent-Source Replacement}
\label{sec:rubric-analysis}

\begin{figure}[t]
 \centering
 \includegraphics[width=1\linewidth]{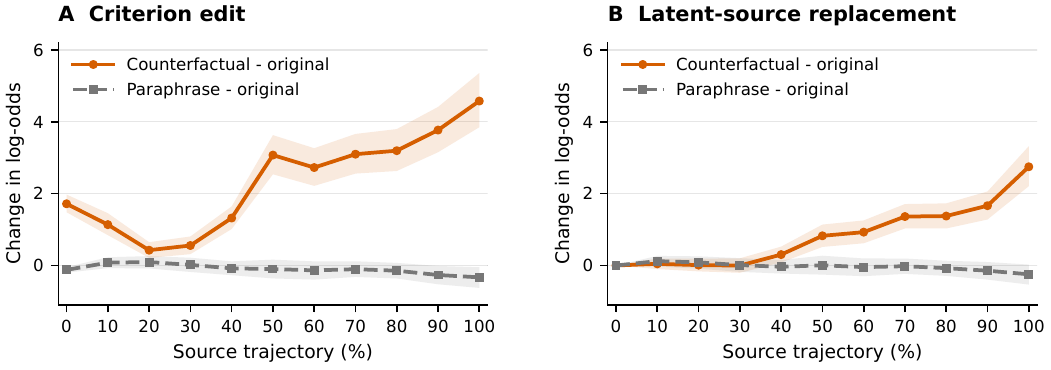}
 \caption{\textbf{Rubric-conditioned preference shifts on 398 pairs.} (A) Each trajectory is generated and evaluated under the same rubric. (B) The evaluation prompt is fixed to the baseline rubric while the latent-trajectory source varies. Changes are measured against the baseline condition; positive values favor the response targeted by the criterion intervention. Bands show pointwise 95\% question-cluster bootstrap intervals.}
 \label{fig:rq1-rubric-preference}
\end{figure}

We test whether changing one evaluation criterion shifts the model's preference and whether the shift is carried by the latent trajectory, motivated by latent-state intervention studies~\citep{li2026latentdynamics}. With all other criteria fixed, the test measures the direction of the preference shift.

\paragraph{Design and readout.}
For each of 398 RewardBench pairs (198 Chat and 200 Chat Hard), we construct three rubrics that differ in one criterion: the \emph{baseline} rubric $\mathcal R_0$ is unchanged; the \emph{wording control} $\mathcal R_p$ paraphrases the selected criterion without changing its meaning; and the \emph{criterion intervention} $\mathcal R_c$ changes that criterion so that it favors a designated response $b_i$. GLM-5.2~\citep{glm5team2026glm5} proposes the two edits, and a fresh-context call verifies that $\mathcal R_p$ preserves meaning and $\mathcal R_c$ changes only the selected criterion, then identifies $b_i$. Neither call receives benchmark labels or judge predictions. We generate trajectories for all three rubrics in both response orders.

At 0\%, 10\%, $\ldots$, 100\% of each trajectory, we probe the preference between the two response labels. For a context $C$ containing the evaluation prompt and latent prefix, we measure the log-odds of the label $\ell_{b_i}$ against the other label $\ell_{\bar b_i}$:
\begin{equation}
 D_i(C)=\log p_\theta(\ell_{b_i}\mid C)-\log p_\theta(\ell_{\bar b_i}\mid C).
 \label{eq:preference-logprob-difference}
\end{equation}
For either edited condition $e\in\{p,c\}$, $\Delta D_i=D_i(C_e)-D_i(C_0)$ measures its effect relative to the baseline; positive values indicate a shift toward $b_i$. This contrastive readout follows~\citet{wang2023ioi,heimersheim2024patching}.

\paragraph{End-to-end rubric intervention.}
The criterion intervention shifts preference toward the designated response, reaching a mean log-odds change of 4.58 at the trajectory endpoint (Figure~\ref{fig:rq1-rubric-preference}A). Its effect becomes substantially larger than that of the wording control in the latter half of the trajectory, indicating sensitivity to the criterion's meaning rather than its phrasing alone. The nonzero shift at 0\% shows a direct effect of the rubric prompt; the replay control below measures the contribution carried by the latent sequence.

\paragraph{Latent-source replacement.}
To remove the visible rubric difference, we fix the evaluation prompt to $\mathcal R_0$ and replay the latent embeddings generated under each rubric in a fresh baseline context. The criterion-intervention source has little effect at early positions, but its influence grows with the supplied trajectory and reaches a mean log-odds change of 2.75 at the endpoint (Figure~\ref{fig:rq1-rubric-preference}B); the wording-control source remains much smaller. Thus, criterion-dependent preference information transfers through the latent sequence even when the evaluation prompt remains unchanged.

\FloatBarrier

\subsection{RQ2: Practical Efficiency of Latent Evaluation}
\label{sec:inference-efficiency}

We evaluate the cost of producing preference judgments. Retrospective interpretation is performed separately and is excluded from these measurements.

\subsubsection{Runtime under Natural Generation}

We compare Rubric-RM-8B and LatentGRM-8B on four domains at vote@5. The vLLM measurements use the same eight-GPU serving configuration and include prefill, scheduling, decoding, and output collection. For reference, we also report full-domain time under ordinary Transformers/PyTorch generation~\citep{wolf2020transformers,paszke2019pytorch} on the same eight GPUs, with one model replica per GPU and five sequential rollouts per assigned sample.

\begin{table}[htbp]
    \centering
    \small
    \setlength{\tabcolsep}{4pt}
    \caption{\textbf{Inference cost at vote@5.} Steps are mean generated positions per rollout. Compression and speedup are the Rubric-RM-8B/LatentGRM-8B ratios of generated steps and total inference time. HF denotes ordinary Transformers/PyTorch generation without vLLM. Bold marks fewer steps or less time and the corresponding compression/speedup ratios within each domain.}
    \label{tab:vote5-efficiency}
    \resizebox{\linewidth}{!}{%
    \begin{tabular}{llrrrrrr}
        \toprule
        \textbf{Domain} ($N$) & \textbf{Model} & \textbf{Steps} & \textbf{Compression} & \textbf{HF time (s)} & \textbf{HF speedup} & \textbf{vLLM time (s)} & \textbf{vLLM speedup} \\
        \midrule
        RB Chat (716) & Rubric-RM-8B & 659.60 & $1.00\times$ & 8196 & $1.00\times$ & 539.70 & $1.00\times$ \\
        & LatentGRM-8B & \textbf{74.38} & $\mathbf{8.87}\times$ & \textbf{1002} & $\mathbf{8.17}\times$ & \textbf{87.76} & $\mathbf{6.15}\times$ \\
        \addlinespace
        RB Chat Hard (912) & Rubric-RM-8B & 582.09 & $1.00\times$ & 11088 & $1.00\times$ & 589.58 & $1.00\times$ \\
        & LatentGRM-8B & \textbf{64.42} & $\mathbf{9.04}\times$ & \textbf{1308} & $\mathbf{8.46}\times$ & \textbf{96.44} & $\mathbf{6.11}\times$ \\
        \addlinespace
        RB2 Precise IF (960) & Rubric-RM-8B & 720.89 & $1.00\times$ & 13416 & $1.00\times$ & 812.93 & $1.00\times$ \\
        & LatentGRM-8B & \textbf{78.59} & $\mathbf{9.17}\times$ & \textbf{1524} & $\mathbf{8.82}\times$ & \textbf{126.99} & $\mathbf{6.40}\times$ \\
        \addlinespace
        RB2 Focus (2970) & Rubric-RM-8B & 695.49 & $1.00\times$ & 37572 & $1.00\times$ & 2350.30 & $1.00\times$ \\
        & LatentGRM-8B & \textbf{75.37} & $\mathbf{9.23}\times$ & \textbf{4476} & $\mathbf{8.39}\times$ & \textbf{338.34} & $\mathbf{6.95}\times$ \\
        \bottomrule
    \end{tabular}}
\end{table}

At vote@5, LatentGRM-8B reduces generated positions by $8.87$--$9.23\times$, yielding $8.17$--$8.82\times$ speedups under ordinary Transformers and $6.11$--$6.95\times$ under vLLM (Table~\ref{tab:vote5-efficiency}). All four domains exceed $6\times$ vLLM speedup despite mixed accuracy differences.

Figure~\ref{fig:accuracy_efficiency_scaling} extends the comparison across vote@1--9. LatentGRM remains faster and scores higher on Chat Hard and Focus at every voting budget; Chat remains below Rubric-RM, and Precise IF varies by budget. More votes generally improve latent judgment quality, with nonmonotonic intermediate results and $6.9\times$--$7.8\times$ speedups at vote@9.

\FloatBarrier
An equal-length control that forces both judges to generate exactly 50 positions yields only a $1.017\times$ runtime difference, showing that latent reasoning does not introduce additional per-step computational cost. (Appendix~\ref{app:fixed50-control}).

\FloatBarrier

\subsection{RQ3: Recovering Rubric-Level Evaluations}
\label{sec:latent-recoverability}
\label{sec:interpreter-results}

Task rubrics provide predefined semantic units for evaluating reconstruction fidelity. We align decoded and teacher evaluations by response and criterion, measuring whether each assessment's state and justification are preserved. This makes changes in individual judgments observable even when most of the reconstructed text remains similar.

Using cached teacher-derived trajectories, we compare three independently trained readouts initialized from the LatentGRM-8B Stage~1 decoder. \textbf{Prompt-only} receives the evaluation input without latent states; \textbf{Replay-hidden} additionally reads projected hidden states obtained by replaying the latent sequence; \textbf{Latent Trace Interpreter (LTI)} reads its vocabulary-probability representation (Section~\ref{sec:latent-interpreter}).

The test set contains 607 records with 9,992 criteria, of which 9,768 have explicit states. State Macro-F1 scores \textsc{Met}, \textsc{Not Met}, and \textsc{Partial}; missing or unparseable states count as errors. Vector Exact requires all states in a fully labeled record to match. Rubric Semantic measures BERTScore similarity~\citep{zhang2020bertscore} between aligned justifications, assigning zero to missing criteria; whole-trace ROUGE-L~\citep{lin2004rouge} provides a lexical comparison.

\begin{table}[htbp]
    \centering
    \small
    \setlength{\tabcolsep}{4pt}
    \caption{\textbf{LatentGRM-8B retrospective reconstruction on the 607-record criterion-rich test set (\%).}}
    \label{tab:interpreter-comparison}
    \resizebox{\linewidth}{!}{%
    \begin{tabular}{lrrrr}
        \toprule
        \textbf{Readout input} & \textbf{State Macro-F1} & \textbf{Vector Exact} & \textbf{Rubric Semantic} & \textbf{ROUGE-L} \\
        \midrule
        Prompt-only & 40.20 & 18.61 & 56.42 & 63.36 \\
        Replay-hidden & 69.42 & 33.03 & 61.29 & 68.22 \\
        Latent Trace Interpreter & \best{98.93} & \best{93.98} & \best{74.65} & \best{79.24} \\
        \bottomrule
    \end{tabular}}
\end{table}

LTI raises State Macro-F1 from 40.20 (Prompt-only) and 69.42 (Replay-hidden) to 98.93, and Vector Exact from 18.61 and 33.03 to 93.98 (Table~\ref{tab:interpreter-comparison}). Prompt-only reaches 63.36 ROUGE-L despite its 40.20 State Macro-F1, showing that lexical overlap misses criterion-state errors. LTI also leads on Rubric Semantic (74.65) and ROUGE-L (79.24), reflecting stronger criterion-level recovery on teacher-derived trajectories.

\paragraph{Downstream utility through data revision.}
We also test whether the recovered evaluation can guide a frozen Qwen3-8B Base reviser on 240 RewardBench Rubric examples. \textbf{Prompt-only} gives the reviser just the question and candidate response; \textbf{Rubric-RM trajectory} supplies explicit textual feedback; \textbf{Latent only} inserts the soft top-10 embedding mixture at each latent step without training the reviser to consume it; and \textbf{Interpreter-restored trajectory} supplies LTI's textual reconstruction of the same latent feedback. The reviser does not see the rubric. Appendix~\ref{app:revision-utility} gives the input construction and scoring protocol.

An external evaluator scores each rubric criterion before and after revision. \textbf{Repair} is the fraction of originally unsatisfied criteria that become satisfied, whereas \textbf{Regression} is the fraction of originally satisfied criteria that become unsatisfied. \textbf{Full} is the percentage of revised answers satisfying every criterion; \textbf{Criterion satisfaction} is the overall percentage of satisfied criterion decisions.

\begin{table}[htbp]
    \centering
    \small
    \setlength{\tabcolsep}{4pt}
    \caption{Frozen-model revision on 240 examples, equally split between Chat and Chat Hard (percent). Repair is computed over 897 originally unsatisfied criteria and Regression over 1,076 originally satisfied criteria. Bold marks the best result in each column.}
    \label{tab:revision-main}
    \begin{tabular}{lrrrr}
        \toprule
        \textbf{Feedback interface} & \textbf{Repair} $\uparrow$ & \textbf{Regression} $\downarrow$ & \textbf{Full} $\uparrow$ & \textbf{Criterion sat.} $\uparrow$ \\
        \midrule
        Prompt-only & 54.07 & 5.20 & 34.17 & 76.28 \\
        Rubric-RM trajectory & 78.71 & 4.65 & 56.67 & 87.78 \\
        Latent only (soft top-10) & 63.32 & 5.11 & 41.67 & 80.54 \\
        Interpreter-restored trajectory & \best{80.49} & \best{1.86} & \best{57.50} & \best{90.12} \\
        \bottomrule
    \end{tabular}
\end{table}

Prompt-only revision reaches 34.17\% Full. Explicit Rubric-RM feedback raises this to 56.67\% and Repair from 54.07\% to 78.71\% (Table~\ref{tab:revision-main}). Even without reviser adaptation, the latent-only interface improves Repair by 9.25 points and Full by 7.50 points over Prompt-only. Interpreter-restored feedback yields the best result in every reported column: 80.49\% Repair, 1.86\% Regression, 57.50\% Full, and 90.12\% criterion satisfaction. Relative to explicit Rubric-RM feedback, its Full rate is comparable while Regression is lower (1.86\% versus 4.65\%).

These results show that latent reasoning preserves the data-synthesis and revision utility of explicit reasoning: after textual restoration, it matches or exceeds the explicit trajectory on every reported metric. More strikingly, the training-free Latent-only interface already improves substantially over Prompt-only revision, demonstrating that latent feedback is directly usable before any specialized adaptation.

\FloatBarrier
\section{Conclusion}
\label{sec:conclusion}

We introduced LatentGRM, a framework for latent reward modeling built on semantic chunking, compression, and reconstruction. Rubric structure guides the compression of explicit evaluations into compact continuous trajectories, which the judge uses to predict pairwise preferences without generating evaluation text. Compared with explicit SFT judges trained on matched data and backbones, LatentGRM achieves competitive aggregate accuracy while substantially reducing trajectory length and inference time. Controlled rubric interventions show that criterion-dependent preference information is carried through the latent sequence, while the Latent Trace Interpreter reconstructs criterion-level assessments for retrospective inspection. Together, these results show how structured latent trajectories can support efficient judgment and make compressed evaluations inspectable.

\bibliography{references}
\bibliographystyle{latentgrm_preprint}

\clearpage
\appendix

\section*{Appendix Contents}
\small
\begin{enumerate}
    \item \hyperref[app:discussion]{Discussion}
    \item \hyperref[app:semantic-chunking]{Semantic Chunking and Exact-Budget Allocation}
    \item \hyperref[app:full-objectives]{Stage~1 Latent Distillation and Stage~2 Autonomous Evaluation}
    \item \hyperref[app:lti-details]{Latent Trace Interpreter Details}
    \item \hyperref[app:rq1-supplement]{Supplementary RQ1: Effectiveness and Rubric-Based Process Evaluation}
    \begin{itemize}
        \setlength{\itemsep}{0pt}
        \setlength{\topsep}{2pt}
        \item Data, backbones, and evaluation protocol
        \item Evaluation-time parameters and Semantic Chunking training dynamics
        \item Rubric-based intervention protocol
        \item Direct judgment without an intermediate evaluation trace
        \item Downstream utility through frozen-model data revision
    \end{itemize}
    \item \hyperref[app:rq2-supplement]{Supplementary RQ2: Practical Efficiency of Latent Evaluation}
    \begin{itemize}
        \setlength{\itemsep}{0pt}
        \setlength{\topsep}{2pt}
        \item vLLM implementation for latent reasoning
        \item Equal-length runtime control
    \end{itemize}
    \item \hyperref[app:interpretability-details]{Supplementary RQ3: Recovering Rubric-Level Evaluations}
    \begin{itemize}
        \setlength{\itemsep}{0pt}
        \setlength{\topsep}{2pt}
        \item Readout data, controls, and evaluation protocol
        \item Metrics, parsing, and reconstruction examples
    \end{itemize}
    \item \hyperref[app:training-setup]{Training Setup and Compute}
    \item \hyperref[app:prompts]{Prompt Templates}
\end{enumerate}
\normalsize
\clearpage

\section{Discussion}
\label{app:discussion}

An important challenge in latent reward modeling is preserving essential evaluation evidence under compression. Fixed-Rate Chunking and the current heuristic Semantic Chunking approximate evaluation structure through fixed token intervals and linguistic boundary cues, respectively. Developing compression strategies that better preserve the evidence most relevant to the final judgment remains an important direction.

A related challenge is recovering the compressed evidence through faithful retrospective interpretation. Appendix~\ref{app:interpreter-mild-failure} illustrates a coherent, nearly complete reconstruction that changes one criterion assessment. The interpreter receives renormalized top-$K$ soft labels, which preserve only a subset of the vocabulary distribution. Richer trajectory representations and reconstruction objectives could strengthen the connection between recovered criterion states and their supporting evidence.

These challenges accompany substantial opportunities. By shortening generation and reducing inference time, latent reasoning can make repeated reward evaluations more affordable in policy learning, data selection, and related workflows. The gains from vote@1 to vote@5 also motivate exploring reinforcement learning to improve individual latent trajectories, building on approaches such as Latent-GRPO~\citep{deng2026latentgrpo}. Together, advances in compression, trajectory optimization, and interpretation could extend the value of latent reward modeling from faster evaluation to more capable and understandable learning systems.

\input{sections/semantic_chunking_appendix}

\begin{figure}[h]
    \centering
    \includegraphics[width=0.72\linewidth]{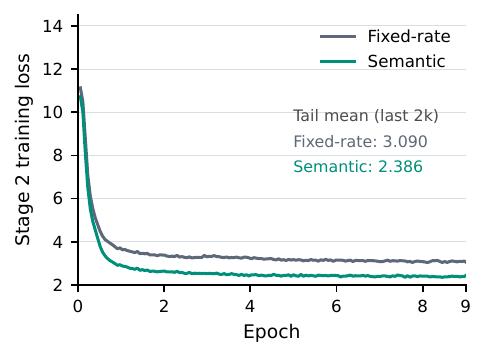}
    \caption{\textbf{Stage~2 loss for the LatentGRM-8B Semantic Chunking ablation.} Semantic Chunking maintains a lower Stage~2 loss after the initial decline.}
    \label{fig:chunking-stage2-loss}
\end{figure}

\section{Stage~1 Latent Distillation and Stage~2 Autonomous Evaluation}
\label{app:full-objectives}

We now give the full objectives for an evaluation input $x$, teacher evaluation $c$, and preference label $y$. Let $\phi$ and $\psi$ denote the trainable Stage~1 encoder and decoder parameters, and let $\theta$ denote the trainable Stage~2 judge parameters. Semantic Chunking specifies $\mathrm{Seg}(c)=\{c_{b_{i-1}+1:b_i}\}_{i=1}^{N}$, adapting the compression frontiers to rubric structure. Throughout this section, $c_{1:M}$ denotes the tokenized evaluation retained under the training sequence-length limit.

\subsection{Encoder Sequence and Vocabulary-Space State}

Let $0=b_0<b_1<\cdots<b_N=M$ be the compression frontiers and let $\kappa_i$ denote the compression placeholder following $c_{b_{i-1}+1:b_i}$. The encoder sequence is
\begin{equation}
\begin{split}
    \xi^{\mathrm{enc}}=[x,\texttt{<think>},
    c_{1:b_1},\kappa_1,c_{b_1+1:b_2},\kappa_2,\ldots,
    c_{b_{N-1}+1:b_N},\kappa_N,\texttt{</think>}].
    \label{eq:appendix-encoder-sequence}
\end{split}
\end{equation}
The encoder attention mask is causal and additionally prevents each query from attending to earlier compression placeholders. If $u$ and $v$ are non-padding query and key positions, respectively, then
\begin{equation}
    A^{\mathrm{enc}}_{uv}=1
    \quad\Longleftrightarrow\quad
    v\leq u\ \land\ \neg(v\in\{\kappa_j\}_{j=1}^N\land v<u).
    \label{eq:encoder-attention-mask}
\end{equation}
Hence $h_i=\operatorname{Enc}_{\phi}(\xi^{\mathrm{enc}};A^{\mathrm{enc}})_{\kappa_i}$ can use $x$ and $c_{1:b_i}$, but neither future evaluation tokens nor earlier compression placeholders.

Let $V$ be the vocabulary size, $d$ the hidden dimension, and $\tau$ the projection temperature. The decoder's frozen output projection $W_{\mathrm{out}}\in\mathbb R^{V\times d}$ maps $h_i$ to vocabulary scores, while its frozen input embedding table $E_{\mathrm{in}}\in\mathbb R^{V\times d}$ maps a vocabulary distribution back to the decoder's input space. The full-vocabulary construction is $\alpha_i=\operatorname{softmax}(W_{\mathrm{out}}h_i/\tau)$ and $z_i^{\mathrm{full}}=E_{\mathrm{in}}^\top\alpha_i$. We retain the indices of the $K$ largest entries in $S_i=\operatorname{TopK}(\alpha_i)$ and compute
\begin{align}
    s_{i,v} &= W_{\mathrm{out}}[v]^{\top}h_i/\tau, \\
    \bar{\alpha}_{i,v} &= \frac{\exp s_{i,v}}
    {\sum_{u\in S_i}\exp s_{i,u}},\qquad v\in S_i, \\
    z_i &= \sum_{v\in S_i}\bar{\alpha}_{i,v}E_{\mathrm{in}}[v].
    \label{eq:appendix-topk-state}
\end{align}
Equivalently, $\bar\alpha_{i,v}=\alpha_{i,v}/\sum_{u\in S_i}\alpha_{i,u}$ on $S_i$, with zero probability outside $S_i$. The full-vocabulary formula is recovered when $K=V$. Gradients pass through the mixture weights to the encoder state $h_i$ but not into $W_{\mathrm{out}}$ or $E_{\mathrm{in}}$.

\subsection{Suffix Reconstruction Loss}

The suffix-reconstruction objective trains each latent prefix to support the subsequent explicit evaluation and final preference. The remaining spans concatenate to $c_{b_i+1:M}$, followed by the preference suffix. Our implementation uniformly samples one frontier and materializes its corresponding causal suffix view.

For frontier $i$, define the decoder prefix and target
\begin{align}
    u_i &= [x,\texttt{<think>},z_{1:i}],\\
    t_i &= [c_{b_i+1:M},\texttt{</think>},y,\texttt{EOS}].
\end{align}
If $L_i$ is the tokenized length of $t_i$, the per-frontier loss is
\begin{equation}
    \ell_i(\phi,\psi)
    =-\frac{1}{L_i}\sum_{j=1}^{L_i}
    \log p_{\psi}\!\left(t_{i,j}\mid u_i,t_{i,<j}\right),
    \label{eq:per-frontier-reconstruction}
\end{equation}
and the complete Stage~1 objective is
\begin{equation}
    \mathcal{L}_{\mathrm{rec}}(\phi,\psi)
    =\frac{1}{N}\sum_{i=1}^{N}\ell_i(\phi,\psi).
    \label{eq:complete-reconstruction}
\end{equation}
The implementation samples $I\sim\operatorname{Uniform}\{1,\ldots,N\}$ for each example and evaluates only $\ell_I$. This estimator is unbiased because
\begin{equation}
    \mathbb{E}_{I}[\ell_I]
    =\sum_{i=1}^{N}\frac{1}{N}\ell_i
    =\mathcal{L}_{\mathrm{rec}}.
    \label{eq:frontier-unbiased-proof}
\end{equation}
For the sampled frontier, $[u_I,t_I]$ uses consecutive position indices and a causal decoder mask. The compressed explicit prefix and future latent positions are omitted, while earlier target tokens remain visible through teacher forcing. Loss is applied only to $t_I$. Each sampled causal view provides one estimate of the full-suffix objective.

\subsection{The Three Stage~1 Optimization Phases}

Let $\phi_0$ and $\psi_0$ denote the initial encoder and decoder parameters. The three phases optimize the same reconstruction objective with different parameter blocks:
\begin{align}
    \widehat{\phi}
    &=\arg\min_{\phi}\mathcal{L}_{\mathrm{rec}}(\phi,\psi_0),
    &&\text{encoder phase: decoder frozen},
    \label{eq:encoder-phase}\\
    \widehat{\psi}
    &=\arg\min_{\psi}\mathcal{L}_{\mathrm{rec}}(\widehat{\phi},\psi),
    &&\text{decoder phase: encoder frozen},
    \label{eq:decoder-phase}\\
    (\phi^*,\psi^*)
    &=\arg\min_{\phi,\psi}\mathcal{L}_{\mathrm{rec}}(\phi,\psi)
    \quad\text{initialized at }(\widehat{\phi},\widehat{\psi}),
    &&\text{joint phase}.
    \label{eq:joint-phase}
\end{align}
After joint training, the encoder runs once over every full teacher trace and exports $(S_i,\bar{\alpha}_i)$ for all frontiers. These sparse targets are stored by dataset index. Stage~2 initializes from the jointly trained decoder adapter, and the encoder's role ends after target export.

\subsection{Stage~2 Soft-Target Training and Inference}

Let $q_\theta$ denote the Stage~2 judge's predicted distribution over the full vocabulary. For every stored top-$K$ target, training resamples independent $g_{i,v}\sim\operatorname{Gumbel}(0,1)$ at each update and forms
\begin{equation}
    \widetilde{\alpha}_{i,v}(\mathbf g)
    =\frac{\exp\left((\log\bar{\alpha}_{i,v}+\lambda_g g_{i,v})/\tau_g\right)}
    {\sum_{u\in S_i}\exp\left((\log\bar{\alpha}_{i,u}+\lambda_g g_{i,u})/\tau_g\right)},
    \quad v\in S_i.
    \label{eq:gumbel-soft-target}
\end{equation}
where $\lambda_g$ controls the noise scale, $\tau_g$ is the perturbation temperature, and $\widetilde\alpha_{i,v}(\mathbf g)=0$ for $v\notin S_i$. This perturbation discourages overfitting to deterministic teacher targets and provides diverse latent trajectories at inference. Teacher forcing inserts
$\widetilde z_i(\mathbf g)=\sum_{v\in S_i}\widetilde\alpha_{i,v}(\mathbf g)E_{\mathrm{in}}[v]$
at the $i$-th latent input position. The prediction immediately preceding this position is matched to the same perturbed distribution, so latent inputs and KL targets share each noise draw. The student distribution remains normalized over the full vocabulary. For one draw, the sparse distillation loss is
\begin{equation}
    \mathcal{L}_{\mathrm{KL}}(\mathbf g)
    =\frac{1}{N}\sum_{i=1}^{N}\sum_{v\in S_i}
    \widetilde\alpha_{i,v}(\mathbf g)
    \log\frac{\widetilde\alpha_{i,v}(\mathbf g)}
    {q_{\theta}(v\mid x,\texttt{<think>},\widetilde z_{<i}(\mathbf g))}.
    \label{eq:sparse-stage2-kl}
\end{equation}
This is teacher-to-student KL: the teacher target is zero outside $S_i$, while the student's normalizer includes all vocabulary entries. Let $r=[\texttt{</think>},y,\texttt{EOS}]$ be the visible suffix. Its cross-entropy is
\begin{equation}
    \mathcal L_{\mathrm{CE}}(r;\mathbf g)
    =-\frac{1}{|r|}\sum_{j=1}^{|r|}
    \log q_\theta(r_j\mid x,\texttt{<think>},\widetilde z_{1:N}(\mathbf g),r_{<j}).
    \label{eq:stage2-explicit-ce}
\end{equation}
Prompt and latent positions are excluded from this CE term. The stochastic objective combines the latent and explicit losses:
\begin{equation}
    \mathcal{L}_{\mathrm{auto}}
    =\mathbb E_{\mathbf g}\!\left[
    \mathcal{L}_{\mathrm{KL}}(\mathbf g)
    +\mathcal{L}_{\mathrm{CE}}(r;\mathbf g)\right].
    \label{eq:full-stage2-loss}
\end{equation}

Both terms depend on the sampled trajectory through teacher-forced inputs. Each training update estimates the expectation with one draw per example.

At inference, the judge generates its own latent trajectory. Given the preceding latent states, step $i$ predicts
\begin{equation}
    q_i=q_\theta(\cdot\mid x,\texttt{<think>},z_{<i}).
    \label{eq:appendix-online-distribution}
\end{equation}
If $\arg\max_v q_i(v)=\texttt{</think>}$, latent generation terminates. Otherwise, we define
\begin{equation}
    S_i^\theta=\operatorname{TopK}(q_i),\qquad
    \bar q_i(v)=
    \begin{cases}
        \displaystyle\frac{q_i(v)}{\sum_{u\in S_i^\theta}q_i(u)}, & v\in S_i^\theta,\\[6pt]
        0, & v\notin S_i^\theta.
    \end{cases}
    \label{eq:appendix-online-topk}
\end{equation}
Replacing $(S_i,\bar\alpha_i)$ in Equation~\ref{eq:gumbel-soft-target} with $(S_i^\theta,\bar q_i)$ yields the perturbed weights $\widetilde q_i(\mathbf g)$. The next latent input is
\begin{equation}
    z_i=\sum_{v\in S_i^\theta}\widetilde q_i(v;\mathbf g)E_{\mathrm{in}}[v].
    \label{eq:appendix-online-feedback}
\end{equation}
The recurrence also ends when the latent-length cap is reached; the judge then feeds the end-marker embedding and greedily decodes the preference label. Distinct Gumbel draws provide the rollout diversity used for voting. Appendix~\ref{app:experimental-details} specifies generation limits, and Appendix~\ref{app:vllm-efficiency} describes the vLLM implementation.

\input{sections/interpreter_appendix}

\section{Supplementary RQ1: Effectiveness and Rubric-Based Process Evaluation}
\label{app:rq1-supplement}
\label{app:experimental-details}

\subsection{Data, Backbones, and Evaluation Protocol}

Each training record contains the pairwise judge prompt $x$, an explicit rubric-based evaluation $c$, and the binary label $y$. The Semantic Chunking cache contains 35,612 OpenRubric SFT records. It contains 27,541,067 explicit evaluation tokens and 3,458,245 latent frontiers, corresponding to 773.36 explicit tokens and 97.11 latent states per example on average. The realized ratio is 7.96 explicit tokens per latent state. These counts are computed after applying the Qwen tokenizer and before Stage~1 frontier subsampling.

We use Qwen3-4B and Qwen3-8B backbones. Stage~1 starts from the base model and uses the dataset's explicit evaluations as supervision.

In the main benchmark runs, a prediction is valid only if its whitespace-trimmed text equals \texttt{Response A} or \texttt{Response B}. Invalid generations cast no vote; ties are resolved by the earliest valid rollout. In the dedicated efficiency runs, an equal vote count or no valid label yields an invalid prediction.

\subsection{Evaluation-Time Parameters and Seeded Diversity}

The main benchmark runs use maximum sequence length 6,144, top-$K=10$, and Gumbel temperature and noise scale of one. Prompts reserve space for the latent start/end markers, 256 latent positions, and eight answer tokens.

Gumbel perturbations supply rollout diversity. For example index $n$ and vote index $r$, the seed is
\begin{equation}
    s_{n,r}=s_{0}+n+r|\mathcal{D}|,
    \qquad s_{0}=42.
    \label{eq:evaluation-gumbel-seed}
\end{equation}
This schedule assigns a distinct noise stream to each example--vote pair and shares seed assignments across voting budgets.

\input{sections/latent_mechanism_appendix}

\input{sections/direct_judge_appendix}

\input{sections/revision_appendix}

\section{Supplementary RQ2: Practical Efficiency of Latent Evaluation}
\label{app:rq2-supplement}

\subsection{vLLM Implementation for Latent Reasoning}
\label{app:vllm-efficiency}

We extend vLLM to pass a continuous embedding back into the decoder at each latent step. The unperturbed argmax determines whether reasoning has ended; otherwise, the next input is a Gumbel-weighted mixture of the top-$K$ token embeddings. After \texttt{</think>}, decoding returns to discrete tokens for the preference label. Each request keeps its own feedback and random state, allowing independent latent rollouts under continuous batching and native parallel sampling.

For tensor-parallel inference, workers select local candidates before forming the global top-$K$ and mix their embeddings on device. Cached input embeddings and persistent feedback buffers support the existing KV cache and CUDA-graph decode path. The explicit CoT baseline uses the same vLLM scheduler and request-submission policy.

\subsection{Equal-Length Runtime Control}
\label{app:fixed50-control}
The 50-position diagnostic covers all 716 Chat and 912 Chat Hard samples at vote@1, using the same prefix-cache reset after warmup. Both samplers set minimum and maximum output length to 50 and ignore EOS; the latent request uses an unreachable latent-end ID to retain soft feedback throughout the measured trajectory. Each model therefore produces exactly 81,400 positions.

\begin{table}[h]
    \centering
    \small
    \caption{\textbf{Equal-length serving throughput.} The full RewardBench Chat and Chat Hard subsets are evaluated at vote@1 with exactly 50 generated positions per sample.}
    \label{tab:fixed50-throughput}
    \resizebox{\linewidth}{!}{%
    \begin{tabular}{lrrrr}
        \toprule
        \textbf{Model} & \textbf{Mean prompt positions} & \textbf{Output positions} & \textbf{Total inference time (s)} & \textbf{Throughput (positions/s)} \\
        \midrule
        Rubric-RM-8B & 913.8 & 81,400 & 46.469 & 1,751.7 \\
        LatentGRM-8B & 914.8 & 81,400 & 45.714 & 1,780.6 \\
        \bottomrule
    \end{tabular}}
\end{table}

LatentGRM adds one prompt position per sample for its start marker (0.11\% of mean prompt length). Total inference time is 46.469 seconds for Rubric-RM-8B and 45.714 seconds for LatentGRM-8B, a difference of only $1.017\times$. Throughput uses the same total-time boundary as natural generation. Forced latent generation never enters visible-answer decoding, so this control measures serving cost without producing accuracy predictions. The near-equal runtime at matched length shows that latent reasoning does not introduce additional per-step computational cost.

\section{Supplementary RQ3: Recovering Rubric-Level Evaluations}
\label{app:interpretability-details}

\subsection{Readout Data and Splits}

The readouts use LatentGRM-8B Stage~1 artifacts and teacher-derived trajectories. The seed-42 split contains 28,430 training, 512 validation, and 3,298 test records.

Of the test pool's 57,097 parsed criteria, 9,768 have explicit states: 7,476 \textsc{Met}, 1,644 \textsc{Not Met}, and 648 \textsc{Partial}. We evaluate all 607 test records containing at least one such state. These records contain 9,992 criteria; 548 are fully labeled and 59 partially labeled.

\subsection{Readout Controls and Evaluation Protocol}

To isolate the information contributed by the saved vocabulary-probability trajectory, we compare LTI with two independently trained controls. All three readouts are initialized from the LatentGRM-8B Stage~1 decoder, keep its pretrained weights frozen, and learn separate LoRA adapters.

\paragraph{Prompt-only.}
This control uses the same reconstruction target as LTI but receives only $[x,\texttt{<think>}]$. Removing the latent positions measures how much of the teacher evaluation can be reconstructed from the prompt alone.

\paragraph{Replay-hidden.}
We replay the latent trajectory through the frozen Stage~2 judge and extract the final-layer state $h_i^{\mathrm{rep}}$ at each latent position. By causal attention, $h_i^{\mathrm{rep}}$ summarizes $x$ and $z_{1:i}$ without access to the explicit evaluation. A learned bridge maps these states into the readout's input space:
\begin{equation}
\begin{split}
    \widehat h_i^{\mathrm{rep}}&=\operatorname{RMSNorm}(h_i^{\mathrm{rep}}),\\
    z_i^{\mathrm{hid}}&=W_{\mathrm{dir}}\widehat h_i^{\mathrm{rep}}+
       W_{\mathrm{up}}\operatorname{SiLU}(W_{\mathrm{down}}\widehat h_i^{\mathrm{rep}}).
\end{split}
    \label{eq:interpreter-hidden-bridge}
\end{equation}
Here $W_{\mathrm{dir}}\in\mathbb R^{d\times d}$ is a direct linear projection, while $W_{\mathrm{down}}\in\mathbb R^{d_b\times d}$ and $W_{\mathrm{up}}\in\mathbb R^{d\times d_b}$ form a nonlinear bottleneck correction. The outputs $z^{\mathrm{hid}}_{1:N}$ replace $z^{\mathrm{int}}_{1:N}$ in Equation~\ref{eq:interpreter-training-sequence}. The bridge and readout LoRA are trained jointly with Equation~\ref{eq:interpreter-full-objective}, without propagating gradients into the Stage~2 judge.

Prompt-only follows the LTI training configuration in Table~\ref{tab:interpreter-settings}. Replay-hidden uses bottleneck width $d_b=512$, learning rate $5\times10^{-6}$, and effective batch size 28; its remaining settings follow LTI. All readouts use eight H20 GPUs. Reconstruction is decoded greedily until \texttt{</think>} or $\min(4096,L_{\mathrm{ref}}+100)$ generated tokens, where $L_{\mathrm{ref}}$ is the reference length. The three outputs are evaluated with the same parser and metrics below.

\subsection{Metrics and Parsing}

State metrics use the explicit labels, Rubric Semantic uses all criterion justifications, and ROUGE-L uses all 607 records.

We parse traces by $g=(\text{response identifier},\text{criterion number},\text{criterion type})$, retaining the last occurrence of duplicate keys. Let $\mathcal G$ contain the 9,768 gold keys with explicit states $y_g\in\{\textsc{Met},\textsc{Not Met},\textsc{Partial}\}$, and let $\hat y_g$ be the parsed prediction. Missing or unparseable predictions count as errors; unlabeled reference criteria are excluded from state scoring.

\paragraph{Criterion State Macro-F1.}
For class $c$, precision and recall are $P_c=TP_c/(TP_c+FP_c)$ and $R_c=TP_c/(TP_c+FN_c)$, with $F1_c=2P_cR_c/(P_c+R_c)$. We report
\begin{equation}
    \operatorname{MacroF1}=\frac{1}{3}\sum_{c\in\{\textsc{Met},\textsc{Not Met},\textsc{Partial}\}}F1_c.
\end{equation}
The three states receive equal weight; missing predictions contribute false negatives.

\paragraph{Criterion Vector Exact.}
Let $\mathcal F$ be the 548 records for which every parsed gold criterion has an explicit state, and let $\mathcal G_i$ be the criterion keys in record $i$. The strict record-level score is
\begin{equation}
    \operatorname{VectorExact}=\frac{1}{|\mathcal F|}\sum_{i\in\mathcal F}
    \mathbf{1}\!\left[\forall g\in\mathcal G_i,\ \hat y_g=y_g\right].
\end{equation}
Partially labeled records contribute to criterion-level metrics only.

\paragraph{Rubric-Aligned Semantic Score.}
For each of the 9,992 parsed gold criteria, we align the predicted and reference justifications by $g$, compute rescaled RoBERTa-large BERTScore-F1 using \texttt{bert-score} 0.3.12, assign zero when the predicted key is missing, and macro-average over gold criteria:
\begin{equation}
    \operatorname{RubricSemantic}=\frac{1}{|\mathcal G_{\rm all}|}
    \sum_{g\in\mathcal G_{\rm all}}\mathbf{1}[g\in\widehat{\mathcal G}]\,
    \operatorname{BERTScoreF1}(\hat e_g,e_g).
\end{equation}
All readouts use the same alignment policy and RoBERTa-large checkpoint. Table~\ref{tab:interpreter-classwise} breaks down criterion-state F1 by class.

\paragraph{ROUGE-L F1.}
We macro-average ROUGE-L F1 between complete generated and reference traces over the 607 records. This measures whole-trace lexical overlap, including shared templates.

\begin{table}[h]
    \centering
    \small
    \setlength{\tabcolsep}{5pt}
    \caption{\textbf{Class-wise criterion-state F1 on the 607-record test set (\%).}}
    \label{tab:interpreter-classwise}
    \begin{tabular}{lrrr}
        \toprule
        \textbf{Readout input} & \textbf{Met F1} & \textbf{Not-Met F1} & \textbf{Partial F1} \\
        \midrule
        Prompt-only & 59.95 & 50.21 & 10.44 \\
        Replay-hidden & 94.45 & 79.46 & 34.36 \\
        Latent Trace Interpreter & \best{99.64} & \best{98.68} & \best{98.46} \\
        \bottomrule
    \end{tabular}
\end{table}

LTI exceeds 98\% F1 for every state (Table~\ref{tab:interpreter-classwise}). The largest separation from Replay-hidden is on \textsc{Partial} (98.46 versus 34.36), showing that LTI better preserves distinctions among criterion judgments.

\subsection{Qualitative Reconstruction Examples}

In test record 14270, LTI recovers all ten criterion states and the comparative rationale. The teacher trace (Figure~\ref{fig:teacher-reference-case-14270}) and LTI reconstruction (Figure~\ref{fig:lti-reconstruction-case-14270}) preserve the decisive facts. Green marks wording in the teacher reference, and purple marks the corresponding semantic paraphrase reconstructed by LTI.

\begingroup
\setlength{\intextsep}{2pt}
\setlength{\abovecaptionskip}{3pt}
\begin{figure}[H]
\centering
\begin{tcolorbox}[enhanced,colback=gray!3!white,colframe=gray!55!black]
\begin{lstlisting}[basicstyle=\ttfamily\scriptsize,breaklines=true,columns=fullflexible,aboveskip=0pt,belowskip=0pt,escapeinside={(*@}{@*)}]
--- Compliance Check ---
Identified Gatekeeper Criterion: Criterion 1: The response must directly address the specific question posed in the request.

--- Analysis ---
**Response A:**
- Criterion 1 [Hard Rule]: Met. The response (*@\textcolor{green!50!black}{directly addresses the calculation requested}@*).
- Criterion 2 [Principle]: Met. The response correctly applies the order of operations (PEMDAS) and performs the calculation accurately.
- Criterion 3 [Principle]: Met. The response (*@\textcolor{green!50!black}{shows the steps: (2-4) = -2, then 2 * -2 = -4}@*).
- Criterion 4 [Principle]: Met. The final answer, -4, is presented explicitly.
- Criterion 5 [Principle]: Met. The response uses standard mathematical notation and the dollar sign for LaTeX formatting.

**Response B:**
- Criterion 1 [Hard Rule]: Met. The response provides an answer to the question.
- Criterion 2 [Principle]: Not Met. The response states the answer is 0, which is incorrect. The correct answer is -4.
- Criterion 3 [Principle]: Not Met. No reasoning or steps are shown.
- Criterion 4 [Principle]: Met. The final answer, 0, is presented explicitly.
- Criterion 5 [Principle]: Met. The response uses standard numerical format.

--- Final Judgment ---
Justification: Response A correctly answers the mathematical question, shows the steps involved, and adheres to all other criteria in the rubric. Response B (*@\textcolor{green!50!black}{fails to provide the correct answer and does not show any reasoning}@*).</think>
\end{lstlisting}
\end{tcolorbox}
\caption{Case study, record 14270: teacher reference evaluation. Green highlights decisive wording in the original trace.}
\label{fig:teacher-reference-case-14270}
\end{figure}

\begin{figure}[H]
\centering
\begin{tcolorbox}[enhanced,colback=blue!2!white,colframe=globalblue]
\begin{lstlisting}[basicstyle=\ttfamily\scriptsize,breaklines=true,columns=fullflexible,aboveskip=0pt,belowskip=0pt,escapeinside={(*@}{@*)}]
--- Compliance Check ---
Identified Gatekeeper Criterion: Criterion 1: The response must directly address the specific question posed in the request.

--- Analysis ---
**Response A:**
- Criterion 1 [Hard Rule]: Met. The response (*@\textcolor{purple!80!black}{directly answers the mathematical question}@*).
- Criterion 2 [Principle]: Met. The response correctly applies the order of operations (PEMDAS).
- Criterion 3 [Principle]: Met. The response (*@\textcolor{purple!80!black}{shows the steps (2-4 = -2, then 2*-2 = -4)}@*).
- Criterion 4 [Principle]: Met. The final answer (-4) is presented explicitly.
- Criterion 5 [Principle]: Met. The response uses appropriate mathematical notation and format.

**Response B:**
- Criterion 1 [Hard Rule]: Met. The response directly answers the mathematical question.
- Criterion 2 [Principle]: Not Met. The response states the answer is 0, which is incorrect. The correct answer is -4.
- Criterion 3 [Principle]: Not Met. No steps or reasoning are shown.
- Criterion 4 [Principle]: Met. The final answer (0) is presented explicitly.
- Criterion 5 [Principle]: Met. The response uses appropriate notation (a number).

--- Final Judgment ---
Justification: Response A correctly answers the mathematical question, shows the correct steps, and adheres to all principles of the rubric. Response B (*@\textcolor{purple!80!black}{provides an incorrect answer and fails to show any reasoning or steps}@*).</think>
\end{lstlisting}
\end{tcolorbox}
\caption{Case study, record 14270: LTI reconstruction. Purple highlights semantic paraphrases of the teacher trace.}
\label{fig:lti-reconstruction-case-14270}
\end{figure}
\endgroup

\subsubsection{A Reconstruction Failure}
\label{app:interpreter-mild-failure}

We additionally examine a reconstruction error from the same frozen 607-record test set. Among complete generations that recover every response--criterion key, record 17234 differs from the teacher reference on exactly one of 16 explicit states. The reconstruction preserves the other states and the final preference for Response~A, but it is locally inconsistent: it marks Response~B's Criterion~3 as \textsc{Met} after noting the correct permutation formula, even though both its later analysis and final judgment recognize that the formula is applied incorrectly. Figures~\ref{fig:mild-failure-teacher-17234} and~\ref{fig:mild-failure-lti-17234} show the teacher trace and reconstruction, respectively; green marks the decisive reference evidence, while red marks the erroneous reconstructed judgment.

\begin{figure}[H]
\centering
\begin{tcolorbox}[enhanced,colback=gray!3!white,colframe=gray!55!black]
\begin{lstlisting}[basicstyle=\ttfamily\scriptsize,breaklines=true,columns=fullflexible,aboveskip=0pt,belowskip=0pt,escapeinside={(*@}{@*)}]
--- Compliance Check ---
Identified Gatekeeper Criterion: Criterion 1: The response must provide a final numerical answer to the question posed. [Hard Rule]

--- Analysis ---
**Response A:**
- Criterion 1 [Hard Rule]: Met. The response provides the numerical answer "30".
- Criterion 2 [Hard Rule]: Met. The response explicitly states "the word MADAM has two A's and two M's" and mentions "five letters".
- Criterion 3 [Hard Rule]: Met. The response correctly applies the formula for permutations with repetitions: 5! / (2! * 2!).
- Criterion 4 [Principle]: Met. The response explains why division is needed (to avoid duplicates) and how the factorial terms are derived (arranging identical letters).
- Criterion 5 [Principle]: Met. The mathematical reasoning for applying the formula and calculating the factorials is consistent.
- Criterion 6 [Principle]: Met. The calculation 5! / (2! * 2!) = 120 / (2 * 2) = 120 / 4 = 30 is accurate.
- Criterion 7 [Principle]: Met. The response clearly states "two A's and two M's, which are indistinguishable from each other" and explains the need to divide by the arrangements of these identical letters.
- Criterion 8 [Principle]: Met. The response follows a logical flow, starting with identifying repetitions, explaining the problem of duplicates, introducing the formula, and then calculating the result.

**Response B:**
- Criterion 1 [Hard Rule]: Met. The response provides the numerical answer "60".
- Criterion 2 [Hard Rule]: Met. The response states "MADAM has five letters, but some of them are repeated. There are two A's and two M's."
- Criterion 3 [Hard Rule]: (*@\colorbox{green!20}{\strut Not Met.}@*) While the response states the correct formula (n! / (r1! r2! ... rk!)), it (*@\textcolor{green!50!black}{incorrectly applies it}@*) by stating the result is 60. The calculation is flawed.
- Criterion 4 [Principle]: Met. The response explains the formula and its components.
- Criterion 5 [Principle]: Not Met. The mathematical reasoning leads to an incorrect conclusion due to a calculation error.
- Criterion 6 [Principle]: Not Met. The calculation (5 times 4 times 3 times 2 times 1) / (2 times 1 times 2 times 1) simplifies to 120 / 4 = 30, not 60. The cancellation step is also described incorrectly ("cancel out some factors... such as 2 and 2, and 1 and 1. This leaves me with (5 times 4 times 3) / 1, which is 60"). This simplification is incorrect.
- Criterion 7 [Principle]: Met. The response identifies the repeated letters.
- Criterion 8 [Principle]: Met. The response presents the reasoning in a step-by-step manner.

--- Final Judgment ---
Justification: Response A correctly identifies the problem, applies the appropriate combinatorial formula for permutations with repetitions, and performs the calculation accurately, arriving at the correct answer of 30. Response B correctly identifies the problem and the formula but makes a significant computational error in its simplification and calculation, leading to an incorrect answer of 60. Therefore, Response A is superior due to its mathematical accuracy.</think>
\end{lstlisting}
\end{tcolorbox}
\caption{Reconstruction failure, record 17234: teacher reference evaluation. Green marks the evidence that Response~B's Criterion~3 is not met.}
\label{fig:mild-failure-teacher-17234}
\end{figure}

\begin{figure}[H]
\centering
\begin{tcolorbox}[enhanced,colback=orange!2!white,colframe=rubricorange]
\begin{lstlisting}[basicstyle=\ttfamily\scriptsize,breaklines=true,columns=fullflexible,aboveskip=0pt,belowskip=0pt,escapeinside={(*@}{@*)}]
--- Compliance Check ---
Identified Gatekeeper Criterion: Criterion 1: The response must provide a final numerical answer to the question posed.

--- Analysis ---
**Response A:**
- Criterion 1 [Hard Rule]: Met. The response provides the numerical answer "30".
- Criterion 2 [Hard Rule]: Met. The response explicitly states "MADAM has two A's and two M's" and identifies them as indistinguishable.
- Criterion 3 [Hard Rule]: Met. The response correctly applies the formula for permutations with repeated elements: 5! / (2! * 2!).
- Criterion 4 [Principle]: Met. The response explains why division by factorials is needed to account for identical elements and how the formula is derived.
- Criterion 5 [Principle]: Met. The mathematical reasoning is accurate and consistent, leading to the correct calculation.
- Criterion 6 [Principle]: Met. The calculation 5! / (2! * 2!) = 120 / 4 = 30 is correct.
- Criterion 7 [Principle]: Met. The response clearly distinguishes between identical (A's and M's) and distinct elements by stating "indistinguishable" and explaining the need to divide by the number of ways to arrange identical letters.
- Criterion 8 [Principle]: Met. The reasoning is presented in a logical, step-by-step manner, starting with identifying repetitions, explaining the need for division, and then performing the calculation.

**Response B:**
- Criterion 1 [Hard Rule]: Met. The response provides the numerical answer "60".
- Criterion 2 [Hard Rule]: Met. The response states "MADAM has five letters, but some of them are repeated. There are two A's and two M's."
- Criterion 3 [Hard Rule]: (*@\colorbox{red!18}{\strut Met.}@*) The response (*@\textcolor{red!70!black}{correctly identifies the formula}@*) for permutations with repeated elements: n! / (r1! r2! ... rk!).
- Criterion 4 [Principle]: Met. The response explains the formula and its components.
- Criterion 5 [Principle]: Not Met. The mathematical reasoning is incorrect. The calculation 5! / (2! * 2!) is 120 / 4 = 30, not 60.
- Criterion 6 [Principle]: Not Met. The calculation (5 * 4 * 3 * 2 * 1) / (2 * 1 * 2 * 1) simplifies to 120 / 4 = 30, not 60. The explanation of canceling factors is also incorrect in this context.
- Criterion 7 [Principle]: Met. The response correctly identifies the repeated elements.
- Criterion 8 [Principle]: Met. The reasoning is presented in a step-by-step manner, although the final calculation is incorrect.

--- Final Judgment ---
Justification: Response A correctly identifies the problem, applies the appropriate combinatorial formula, and performs the calculation accurately, arriving at the correct answer of 30. Response B correctly identifies the formula but makes a significant computational error in the final calculation, leading to an incorrect answer of 60. Therefore, Response A is superior due to its accuracy.</think>
\end{lstlisting}
\end{tcolorbox}
\caption{Reconstruction failure, record 17234: LTI reconstruction. Red marks the erroneous \textsc{Met} state for Response~B's Criterion~3.}
\label{fig:mild-failure-lti-17234}
\end{figure}

\clearpage
\section{Training Setup and Compute}
\label{app:training-setup}

We train separate Qwen3-4B and Qwen3-8B LatentGRM models on the same 35,612 OpenRubrics records. Each run uses one node with eight NVIDIA RTX 6000D GPUs (85,651 MiB of visible memory per GPU). Training uses bfloat16 precision, DeepSpeed ZeRO Stage~2, gradient checkpointing, and LoRA. Stage~1 uses SDPA for its four-dimensional encoder attention mask, whereas Stage~2 uses FlashAttention-2. Table~\ref{tab:training-hyperparameters} summarizes the remaining settings.

\begin{table}[h]
    \centering
    \small
    \caption{Training configuration for LatentGRM. Effective batch size includes eight data-parallel workers and gradient accumulation.}
    \label{tab:training-hyperparameters}
    \begin{tabular}{p{0.34\linewidth}p{0.58\linewidth}}
        \hline
        \textbf{Setting} & \textbf{Value} \\
        \hline
        Backbones / training records & Qwen3-4B and Qwen3-8B; 35,612 records \\
        Hardware per run & $8\times$ NVIDIA RTX 6000D (85,651 MiB per GPU) \\
        Precision / distributed training & bfloat16; DeepSpeed ZeRO Stage~2; gradient checkpointing \\
        Maximum sequence length & 6,144 tokens \\
        Compression rate / latent support & $\rho=8$; top-$K=10$; projection temperature $\tau=1$ \\
        Semantic Chunking constraints & hard 4--16 tokens; preferred 6--10 tokens; local search radius 4 tokens (5 for lexical rescue) \\
        Stage~1 encoder/decoder/joint phases & 3 epochs each; learning rate $3\times10^{-5}$ \\
        Stage~1 batch configuration & 1 per device; 8 accumulation steps; effective batch 64 \\
        Stage~2 reported checkpoint & Epoch 9; learning rate $1\times10^{-5}$ \\
        Stage~2 batch configuration & 2 per device; 16 accumulation steps; effective batch 256 \\
        Optimizer and schedule & AdamW; weight decay 0.01; 5\% warmup; WarmupDecayLR \\
        LoRA & rank 64; scaling $\alpha=128$; dropout 0.1; no bias \\
        Objective weights & $\lambda_{\mathrm{KL}}=1$ and $\lambda_{\mathrm{pref}}=1$ \\
        Training Gumbel perturbation & temperature 1.0; noise scale 1.0 \\
        Random seed & 42 \\
        Checkpointing / logging & one checkpoint per epoch; log every 10 optimizer steps \\
        \hline
    \end{tabular}
\end{table}

The encoder LoRA targets the attention projections, MLP projections, and \texttt{embed\_tokens}; the last module learns the newly introduced \texttt{<|compress\_token|>} embedding. Decoder and Stage~2 adapters target the attention and MLP projections. The vocabulary output matrix and input embedding basis are accessed separately, which is necessary for the untied Qwen3-8B backbone. Soft labels are generated in bfloat16 with batch size 16 and stored in shards of 1,000 examples. Deterministic tokenization, Semantic Chunking boundaries, and frontier indices are cached once; only the uniformly sampled Stage~1 supervision frontier remains dynamic across epochs.

\clearpage
\section{Prompt Templates}
\label{app:prompts}

\begin{figure}[htbp]
\centering
\begin{tcblisting}{
    listing only,
    listing engine=listings,
    colback=gray!5!white,
    colframe=gray!75!black,
    title=Rubric Generation Prompt,
    fonttitle=\bfseries,
    boxrule=0.6pt,
    arc=1mm,
    left=1mm,
    right=1mm,
    top=1mm,
    bottom=1mm,
    listing options={
        basicstyle=\ttfamily\footnotesize,
        breaklines=true,
        breakatwhitespace=true,
        columns=fullflexible,
        keepspaces=true,
        showstringspaces=false
    }
}
Your task is to extract a set of rubric-style instructions from a user's request. These rubrics will be used as evaluation criteria to check if a response fully meets the request.

Rubric items should be broadly reusable whenever possible. Topic-specific details (e.g., names, places, numbers, or required content) are allowed in a [Hard Rule] only when they are necessary to preserve an explicit requirement from the request. Unnecessary topic-specific details are invalid.

- **Two Distinct Categories:**
  - [Hard Rule]: Preserve explicit requirements stated in the <request> exactly (format, length, structure, names, numbers, forbidden/required elements, etc.).
  - [Principle]: Derived by abstracting any concrete cues into domain-agnostic quality criteria (e.g., clarity, correctness, sound reasoning, pedagogy).
- **Comprehensiveness:** The rubric must cover all critical aspects implied by the request (including any examples contained in it), spanning explicit requirements and implicit quality standards.
- **Conciseness & Uniqueness:** Each rubric must capture a distinct evaluation criterion. Overlapping or redundant criteria must be merged into a single rubric. Wording must be precise and free of repetition.
- **Format Requirements:**
  - Use a numbered list.
  - Each item starts with "The response" phrased in third person.
  - Append [Hard Rule] or [Principle] at the end of each item.
  - Do not include reasoning, explanations, or examples in the final output, only the rubrics.

Here is the request:
{instruction}

Please generate the rubrics for the above request.
\end{tcblisting}
\caption{Prompt used to construct task-specific rubric criteria.}
\label{fig:rubric-generation-prompt}
\end{figure}

\begin{figure}[htbp]
\centering
\begin{tcblisting}{
    listing only,
    listing engine=listings,
    colback=gray!5!white,
    colframe=gray!75!black,
    title=Explicit Evaluation-Trace Prompt,
    fonttitle=\bfseries,
    boxrule=0.6pt,
    arc=1mm,
    left=1mm,
    right=1mm,
    top=1mm,
    bottom=1mm,
    listing options={
        basicstyle=\ttfamily\scriptsize,
        breaklines=true,
        breakatwhitespace=true,
        columns=fullflexible,
        keepspaces=true,
        showstringspaces=false
    }
}
You are a fair and impartial judge. Your task is to evaluate 'Response A' and 'Response B' based on a given instruction and a rubric. You will conduct this evaluation in distinct phases as outlined below.

### Phase 1: Compliance Check Instructions
First, identify the single most important, objective 'Gatekeeper Criterion' from the rubric.
- **A rule is objective (and likely a Gatekeeper) if it can be verified without opinion. Key examples are: word/paragraph limits, required output format (e.g., JSON validity), required/forbidden sections, or forbidden content.**
- **Conversely, a rule is subjective if it requires interpretation or qualitative judgment. Subjective rules about quality are NOT Gatekeepers. Examples include criteria like "be creative," "write clearly," "be engaging," or "use a professional tone."**
Think step-by-step to determine this single most important Gatekeeper.

### Phase 2: Analyze Each Response
Next, for each Gatekeeper Criterion and all other criteria in the rubric, evaluate each response item by item. For each item, think step-by-step and cite concrete evidence from the response before assigning your judgment.

### Phase 3: Final Judgment Instructions
Based on the results from the previous phases, determine the winner using these simple rules. Provide a final justification explaining your decision first and then give your decision. Think step-by-step to aggregate the findings and make the decision; keep the reasoning explicit and concise.

---
### REQUIRED OUTPUT FORMAT
You must follow this exact output format below.
--- Compliance Check ---
Identified Gatekeeper Criterion: <e.g., Criterion 1: Must be under 50 words.>

--- Analysis ---
**Response A:**
- Criterion 1 [Hard Rule]: Justification: <...>
- Criterion 2 [Hard Rule]: Justification: <...>
- Criterion 3 [Principle]: Justification: <...>
- ... (and so on for all other criteria)

**Response B:**
- Criterion 1 [Hard Rule]: Justification: <...>
- Criterion 2 [Hard Rule]: Justification: <...>
- Criterion 3 [Principle]: Justification: <...>
- ... (and so on for all other criteria)

--- Final Judgment ---
Justification: <...>
Winner: <Response A / Response B>

Task to Evaluate:
Instruction:
{instruction}
Rubric:
{rubric}
Response A:
{response_a}
Response B:
{response_b}
\end{tcblisting}
\caption{Prompt used to elicit a structured explicit evaluation trace.}
\label{fig:explicit-trace-prompt}
\end{figure}

\begin{figure}[htbp]
\centering
\begin{tcblisting}{
    listing only,
    listing engine=listings,
    colback=gray!5!white,
    colframe=gray!75!black,
    title=LatentGRM Training and Evaluation Prompt,
    fonttitle=\bfseries,
    boxrule=0.6pt,
    arc=1mm,
    left=1mm,
    right=1mm,
    top=1mm,
    bottom=1mm,
    listing options={
        basicstyle=\ttfamily\footnotesize,
        breaklines=true,
        breakatwhitespace=true,
        columns=fullflexible,
        keepspaces=true,
        showstringspaces=false
    }
}
You are a fair and impartial judge. Your task is to evaluate 'Response A' and 'Response B' based on a given instruction and a rubric. You will conduct this evaluation in distinct phases as outlined below.

### Phase 1: Compliance Check Instructions
First, internally identify the single most important, objective 'Gatekeeper Criterion' from the rubric.
- **A rule is objective (and likely a Gatekeeper) if it can be verified without opinion. Key examples are: word/paragraph limits, required output format (e.g., JSON validity), required/forbidden sections, or forbidden content.**
- **Conversely, a rule is subjective if it requires interpretation or qualitative judgment. Subjective rules about quality are NOT Gatekeepers. Examples include criteria like "be creative," "write clearly," "be engaging," or "use a professional tone."**
Reason step-by-step internally to determine this single most important Gatekeeper.

### Phase 2: Analyze Each Response
Internally evaluate Response A and Response B against the Gatekeeper Criterion and every other criterion in the rubric. For each criterion, reason step-by-step and use concrete evidence from the responses.

### Phase 3: Final Judgment Instructions
Internally aggregate the findings and determine the better response. Keep the reasoning precise and consistent with the rubric.

### REQUIRED OUTPUT FORMAT
Output exactly one label and nothing else: Response A or Response B.

Task to Evaluate:
Instruction:
{instruction}
Rubric:
{rubric}
Response A:
{response_a}
Response B:
{response_b}
\end{tcblisting}
\caption{Prompt used for LatentGRM training and inference.}
\label{fig:latentgrm-prompt}
\end{figure}

\end{document}

%% file: math_commands.tex
\usepackage{amsmath,amsfonts,bm}

\def\eqref#1{equation~\ref{#1}}

\def\1{\bm{1}}

\DeclareMathAlphabet{\mathsfit}{\encodingdefault}{\sfdefault}{m}{sl}
\SetMathAlphabet{\mathsfit}{bold}{\encodingdefault}{\sfdefault}{bx}{n}



%% file: figures/semantic_chunking_illustration.tex
\begin{figure}[t]
\centering
\begin{tcolorbox}[
    enhanced, colback=gray!3!white, colframe=gray!60!black,
    colbacktitle=gray!12!white, coltitle=black,
    title=\textbf{Example 1: preserving rubric structure and complete claims},
    fonttitle=\small, boxrule=0.55pt, arc=1mm,
    left=1mm, right=1mm, top=0.5mm, bottom=0.5mm]
\begin{minipage}[t]{0.485\linewidth}
\begin{tcolorbox}[
    enhanced, equal height group=seg-example-one,
    colback=semanticgreen!4!white, colframe=semanticgreen!75!black,
    colbacktitle=semanticgreen!12!white, coltitle=semanticgreen!45!black,
    title={\centering Semantic Chunking}, fonttitle=\small,
    boxrule=0.7pt, arc=1mm,
    left=1mm, right=1mm, top=0.5mm, bottom=0.5mm]
\footnotesize\raggedright
- Criterion 4 [Principle]: Met. \semanticbar{} The logic for generating the metaphor is clear. \semanticbar{} It randomly selects words and combines them into a sentence.
\end{tcolorbox}
\end{minipage}\hfill
\begin{minipage}[t]{0.485\linewidth}
\begin{tcolorbox}[
    enhanced, equal height group=seg-example-one,
    colback=fixedred!4!white, colframe=fixedred!75!black,
    colbacktitle=fixedred!12!white, coltitle=fixedred!55!black,
    title={\centering Fixed-Rate Chunking ($\rho=8$)}, fonttitle=\small,
    boxrule=0.7pt, arc=1mm,
    left=1mm, right=1mm, top=0.5mm, bottom=0.5mm]
\footnotesize\raggedright
- Criterion 4 [Principle]: \fixedbar{} Met. The logic for generating the metaphor \fixedbar{} is clear. It randomly selects words and \fixedbar{} combines them into a sentence.
\end{tcolorbox}
\end{minipage}
\end{tcolorbox}

\vspace{0.2em}

\begin{tcolorbox}[
    enhanced, colback=gray!3!white, colframe=gray!60!black,
    colbacktitle=gray!12!white, coltitle=black,
    title=\textbf{Example 2: retaining sentence-level evidence},
    fonttitle=\small, boxrule=0.55pt, arc=1mm,
    left=1mm, right=1mm, top=0.5mm, bottom=0.5mm]
\begin{minipage}[t]{0.485\linewidth}
\begin{tcolorbox}[
    enhanced, equal height group=seg-example-two,
    colback=semanticgreen!4!white, colframe=semanticgreen!75!black,
    colbacktitle=semanticgreen!12!white, coltitle=semanticgreen!45!black,
    title={\centering Semantic Chunking}, fonttitle=\small,
    boxrule=0.7pt, arc=1mm,
    left=1mm, right=1mm, top=0.5mm, bottom=0.5mm]
\footnotesize\raggedright
It does not provide an explanation or a flowchart. \semanticbar{} Response B correctly explains the code and its logic.
\end{tcolorbox}
\end{minipage}\hfill
\begin{minipage}[t]{0.485\linewidth}
\begin{tcolorbox}[
    enhanced, equal height group=seg-example-two,
    colback=fixedred!4!white, colframe=fixedred!75!black,
    colbacktitle=fixedred!12!white, coltitle=fixedred!55!black,
    title={\centering Fixed-Rate Chunking ($\rho=8$)}, fonttitle=\small,
    boxrule=0.7pt, arc=1mm,
    left=1mm, right=1mm, top=0.5mm, bottom=0.5mm]
\footnotesize\raggedright
It does not provide an explanation or \fixedbar{} a flowchart. Response B correctly explains \fixedbar{} the code and its logic.
\end{tcolorbox}
\end{minipage}
\end{tcolorbox}
\caption{\textbf{Illustrative frontiers in cropped evaluations.} Green and red mark the semantic and fixed-rate cuts on identical text.}
\label{fig:segmentation-examples}
\end{figure}

%% file: sections/semantic_chunking_appendix.tex
\section{Semantic Chunking and Exact-Budget Allocation}
\label{app:semantic-chunking}

Semantic Chunking first divides an evaluation into outer rubric blocks and allocates a fixed global latent budget across them. It then refines each block's internal boundaries while holding its allocated count fixed. The procedure preserves the original token sequence and produces exactly $N=\lceil M/8\rceil$ latent positions. Preprocessing is performed offline and cached. Figures~\ref{fig:complete-three-region-chunking} and~\ref{fig:complete-fixed-rate-chunking} illustrate Semantic Chunking and Fixed-Rate Chunking.

\subsection{Rubric Structure and Initial Budget Allocation}

We tokenize the evaluation once with character offsets. The outer blocks comprise the initial compliance and gatekeeper discussion, each response's individual criterion assessments, and the final synthesis; each response header joins its first criterion. Structural expressions of at most 10 tokens are protected against internal cuts, including section and response headers, criterion identifiers, \texttt{[Hard Rule]}, \texttt{[Principle]}, \texttt{Justification:}, \texttt{Met.}, and \texttt{Not Met.} We use \emph{boundary} for a selected token position in the partition; each selected boundary serves as a compression frontier in Stage~1.

Let $m_s$ be the token length of outer block $s$, for $s=1,\ldots,S$, and let $n_s$ be its latent budget. We require
\begin{equation}
    \sum_{s=1}^{S}n_s=N,\qquad
    \left\lceil\frac{m_s}{16}\right\rceil\leq n_s
    \leq\left\lfloor\frac{m_s}{4}\right\rfloor,
    \qquad N=\left\lceil\frac{\sum_s m_s}{8}\right\rceil.
    \label{eq:appendix-budget-constraints}
\end{equation}
For a candidate count $k$, let $d_{s,1:k}$ be the balanced integer partition of $m_s$, with lengths differing by at most one. Define $\delta(d)=\max(6-d,0,d-10)$. The allocation cost is
\begin{equation}
    A_s(k)=\left(
        \sum_j\mathbf{1}[\delta(d_{s,j})>0],\quad
        \sum_j\delta(d_{s,j}),\quad
        \sum_j(d_{s,j}-8)^2
    \right).
    \label{eq:semantic-allocation-cost}
\end{equation}
We minimize the sum of these costs lexicographically using
\begin{equation}
    D(s,n)=\min_k^{\mathrm{lex}}\{D(s-1,n-k)+A_s(k)\},
    \qquad D(0,0)=(0,0,0).
    \label{eq:global-budget-dp}
\end{equation}
The three priorities are the number of lengths outside 6--10 tokens, their distance from this interval, and their deviation from the target rate.

With $n_s$ fixed, a local dynamic program constructs feasible reference boundaries $b^0_{s,1:n_s}$ using lengths in $[4,16]$ and avoiding protected expressions. It applies the same three priorities to the realized segment lengths, then uses total displacement from the evenly spaced boundaries $\operatorname{round}(j m_s/n_s)$ as the final tie-breaker.

\subsection{Linguistic Boundary Costs}

We combine rule-based sentence detection with a spaCy dependency parser to identify natural sentence and clause boundaries. Before parsing, structural templates, code, mathematics, and markup are replaced with equal-length whitespace. This prevents non-prose spans from interfering with the syntactic analysis while preserving character alignment, so linguistic evidence can be mapped back to the original token boundaries.

Each candidate boundary receives a base cost $\beta(b)$ from local linguistic evidence. We prefer boundaries at the ends of sentences or complete clauses, and discourage boundaries that split structural labels, attached punctuation, compact code or mathematical expressions, or lexical units. Sentence endings are detected with deterministic rules that avoid common false positives such as decimals, ellipses, abbreviations, and protected expressions. Table~\ref{tab:semantic-boundary-costs} summarizes these preferences.

The dependency parse additionally assigns a local risk level $r(b)\in\{0,1,2\}$, where higher values indicate a more severe disruption of a short local dependency. We consider only relations spanning at most eight tokens so that long parser arcs do not penalize large portions of a sentence. The combined boundary cost is
\begin{equation}
    C(b)=\beta(b)+3r(b).
    \label{eq:semantic-boundary-cost}
\end{equation}
Cuts inside words, numbers, or identifiers receive a high base cost and are also marked separately, allowing lexical integrity to be optimized before the remaining soft preferences. We additionally use a rule-based fragment penalty $F(a,b)\in\{0,4,8\}$ to discourage short, incomplete rubric fragments, while assigning zero cost to complete headers and status statements.

\begin{table}[H]
    \centering
    \small
    \caption{Base costs for candidate internal boundaries. Lower costs favor a cut; protected expressions are excluded before scoring.}
    \label{tab:semantic-boundary-costs}
    \begin{tabular}{p{0.76\linewidth}r}
        \toprule
        \textbf{Boundary type} & \textbf{Base cost} $\beta(b)$ \\
        \midrule
        Observed sentence end & $-3$ \\
        Complete criterion-header end or predicate-bearing clause edge & $0.5$ \\
        Soft punctuation, such as a comma, colon, or semicolon & $1$ \\
        Lexical separator within code or mathematics & $1.5$ \\
        Ordinary word boundary & $2$ \\
        Internal position of a criterion header & $6$ \\
        Boundary that detaches prose punctuation or quotation delimiters & $8$ \\
        Internal position of a short literal, word, number, or identifier & $12$ \\
        \bottomrule
    \end{tabular}
\end{table}

\subsection{Constrained Local Refinement}

Consider one outer block of length $m$ with fixed count $n$ and reference boundaries $b^0_{1:n}$. A candidate partition $b_{0:n}$ has $b_0=0$, $b_n=m$, and lengths $d_j=b_j-b_{j-1}$. We require
\begin{equation}
\begin{split}
    &4\leq d_j\leq16,\qquad b_j\notin\mathcal F\quad(j<n),\\
    &|b_j-b_j^0|\leq w,\qquad
    \sum_{j<n}\mathbf{1}[r(b_j)=2]
    \leq\sum_{j<n}\mathbf{1}[r(b_j^0)=2],
\end{split}
    \label{eq:semantic-refinement-constraints}
\end{equation}
where $\mathcal F$ contains the protected boundaries and the default search radius is $w=4$ tokens.

Let $W(\mathbf b)$ count the internal boundaries that cut a lexical unit. Among feasible partitions, we first minimize $W$, then minimize
\begin{equation}
\begin{split}
    J(\mathbf b)={}&\sum_{j<n}C(b_j)+\sum_{j=1}^{n}F(b_{j-1},b_j)
       +0.50\sum_{j=1}^{n}\delta(d_j)^2\\
       &+0.02\sum_{j=1}^{n}(d_j-m/n)^2
       +0.10\sum_{j<n}|b_j-b_j^0|.
\end{split}
    \label{eq:semantic-refinement-objective}
\end{equation}
The first two terms favor linguistic boundaries and complete fragments; the remaining terms control preferred length, length variance, and displacement from the reference partition.

A dynamic-programming state records the chunk count, current boundary, and risk-2 count. Transitions add feasible chunks and minimize $(W,J)$ lexicographically, breaking ties by the number of moved boundaries and then their coordinates.

We adopt the candidate if it reduces $W$. When $W$ is unchanged, we require $J$ not to increase and the boundary-plus-fragment cost
$H(\mathbf b)=\sum_{j<n}C(b_j)+\sum_jF(b_{j-1},b_j)$
to decrease by at least 0.5; otherwise we retain the reference.

\begin{algorithm}[htbp]
\caption{Syntax-Aware Semantic Chunking with an Exact Latent Budget}
\label{alg:semantic-plan}
\small
\DontPrintSemicolon
\SetAlgoNoEnd
\SetAlgoNlRelativeSize{-1}
\SetKwInOut{KwRequire}{Require}
\SetKwInOut{KwReturn}{Return}
\KwRequire{teacher evaluation $c$, tokenizer $\mathcal T$, rate $\rho=8$}
$(\mathbf v,\mathbf o)\leftarrow\textsc{TokenizeOnce}(c,\mathcal T)$\;
$\mathcal O\leftarrow\textsc{OuterRubricBlocks}(c,\mathbf o)$\;
$N\leftarrow\lceil|\mathbf v|/8\rceil$\;
$(n_1,\ldots,n_S)\leftarrow\textsc{AllocateBudget}(\mathcal O,N)$\;
$\mathcal F\leftarrow\textsc{ProtectedStructuralCuts}(c,\mathbf o)$\;
$(C,r,W,F)\leftarrow\textsc{BoundaryAndFragmentEvidence}(c,\mathbf o)$\;
\ForEach{outer block $s$}{
    $\mathbf b_s^0\leftarrow\textsc{ProtectedReferencePartition}(m_s,n_s,\mathcal F_s)$\;
    $\mathbf b_s\leftarrow\textsc{RefineAndSelect}(\mathbf b_s^0,C,r,W,F,w=4)$\;
    \If{$W(\mathbf b_s)>0$}{
        $\mathbf b_s'\leftarrow\textsc{RefineAndSelect}(\mathbf b_s^0,C,r,W,F,w=5)$\;
        \If{$W(\mathbf b_s')<W(\mathbf b_s)$}{$\mathbf b_s\leftarrow\mathbf b_s'$\;}
    }
}
\KwReturn{concatenated block-local boundaries, shifted to global token offsets}
\end{algorithm}

Refinement leaves each outer block's span and latent count unchanged without increasing lexical or risk-2 cuts. Stage~1 uses the resulting compression frontiers with the supervision objective in Appendix~\ref{app:full-objectives}.

\FloatBarrier
\subsection{Realized Allocation and a Complete Example}

The training cache contains 35,612 evaluations, 27,541,067 explicit tokens, and 3,458,245 latent positions: 97.11 latents per evaluation and 7.96 explicit tokens per latent. The per-example latent counts match Fixed-Rate Chunking. As Table~\ref{tab:semantic-length-distribution} shows, 86.18\% of chunks contain 6--10 tokens. Refinement moves 1,836,211 internal boundaries, reducing lexical cuts from 142,295 to 151 and risk-2 cuts from 694,857 to 946.

\begin{table}[htbp]
    \centering
    \small
    \caption{Chunk lengths in the Qwen3-8B semantic training cache. All 35,612 evaluations retain their exact rate-8 latent budgets.}
    \label{tab:semantic-length-distribution}
    \begin{tabular}{lrr}
        \toprule
        \textbf{Explicit tokens per latent} & \textbf{Chunk count} & \textbf{Share (\%)} \\
        \midrule
        4 & 37,514 & 1.08 \\
        5 & 223,876 & 6.47 \\
        6 & 399,275 & 11.55 \\
        7 & 572,989 & 16.57 \\
        8 & 1,053,395 & 30.46 \\
        9 & 523,161 & 15.13 \\
        10 & 431,511 & 12.48 \\
        11 & 185,989 & 5.38 \\
        12 & 28,541 & 0.83 \\
        13--16 & 1,994 & 0.06 \\
        \midrule
        Total & 3,458,245 & 100.00 \\
        \bottomrule
    \end{tabular}
\end{table}

Figures~\ref{fig:complete-three-region-chunking} and~\ref{fig:complete-fixed-rate-chunking} compare the boundary locations produced by Semantic Chunking and Fixed-Rate Chunking on the same cached evaluation, containing 219 explicit tokens and 28 latent positions.

\input{sections/semantic_chunking_examples}
\FloatBarrier

%% file: sections/semantic_chunking_examples.tex
\begin{figure}[htbp]
\centering
\begin{tcolorbox}[
    enhanced,
    colback=white,
    colframe=gray!65!black,
    boxrule=0.7pt,
    arc=1mm,
    boxsep=0pt,
    left=0pt,
    right=0pt,
    top=0pt,
    bottom=0pt
]
\begin{tcolorbox}[
    enhanced,
    frame hidden,
    sharp corners,
    colback=globalblue!7!white,
    boxrule=0pt,
    boxsep=0pt,
    left=1.2mm,
    right=1.2mm,
    top=1mm,
    bottom=1mm,
    before skip=0pt,
    after skip=0pt,
    borderline south={0.45pt}{0pt}{gray!55!white}
]
\begin{minipage}[c]{0.76\linewidth}
{\ttfamily\scriptsize\raggedright
--- Compliance Check ---\\
Identified Gatekeeper Criterion:\examplebar{} Criterion 1: The response directly addresses\examplebar{} the question posed by providing\examplebar{} a clear and explicit answer.\\
--- Analysis ---\examplebar{}\\

}\end{minipage}\hfill
\begin{minipage}[c]{0.20\linewidth}
\raggedleft\color{globalblue!55!black}\bfseries\footnotesize
Global assessment /\\[-0.1em]gatekeeper discussion
\end{minipage}
\end{tcolorbox}
\begin{tcolorbox}[
    enhanced,
    frame hidden,
    sharp corners,
    colback=rubricorange!7!white,
    boxrule=0pt,
    boxsep=0pt,
    left=1.2mm,
    right=1.2mm,
    top=1mm,
    bottom=1mm,
    before skip=0pt,
    after skip=0pt,
    borderline south={0.45pt}{0pt}{gray!55!white}
]
\begin{minipage}[c]{0.76\linewidth}
{\ttfamily\scriptsize\raggedright
**Response A:**\examplebar{}\\
- Criterion 1 [Hard Rule]: Met.\examplebar{} The response provides the explicit answer "7".\examplebar{}\\
- Criterion 2 [Principle]: Met.\examplebar{} The answer "7"\examplebar{} is self-contained and clear.\examplebar{}\\
- Criterion 3 [Principle]: Met.\examplebar{} The response uses conventional notation\examplebar{} (a numeral) to communicate the answer.\examplebar{}\\
**Response B:**\examplebar{}\\
- Criterion 1 [Hard Rule]: Met.\examplebar{} The response provides the explicit answer "7".\examplebar{}\\
- Criterion 2 [Principle]: Met.\examplebar{} The answer is presented in a complete sentence,\examplebar{} making it self-contained and clear.\examplebar{}\\
- Criterion 3 [Principle]: Met.\examplebar{} The response uses appropriate and conventional\examplebar{} language to communicate the answer effectively.\examplebar{}\\

}\end{minipage}\hfill
\begin{minipage}[c]{0.20\linewidth}
\raggedleft\color{rubricorange!55!black}\bfseries\footnotesize
Response-wise\\[-0.1em]rubric analysis
\end{minipage}
\end{tcolorbox}
\begin{tcolorbox}[
    enhanced,
    frame hidden,
    sharp corners,
    colback=finalpurple!7!white,
    boxrule=0pt,
    boxsep=0pt,
    left=1.2mm,
    right=1.2mm,
    top=1mm,
    bottom=1mm,
    before skip=0pt,
    after skip=0pt
]
\begin{minipage}[c]{0.76\linewidth}
{\ttfamily\scriptsize\raggedright
--- Final Judgment ---\\
Justification:\examplebar{} Both responses successfully meet all criteria.\examplebar{} Response B is slightly better as it provides\examplebar{} a more complete and conversational answer,\examplebar{} directly addressing the question in a full sentence\examplebar{} while still being concise and clear.\examplebar{} \\[0.35em]
\textbf{Preference label (outside the chunked trace): Response B}

}\end{minipage}\hfill
\begin{minipage}[c]{0.20\linewidth}
\raggedleft\color{finalpurple!55!black}\bfseries\footnotesize
Final synthesis\\[-0.1em]and judgment
\end{minipage}
\end{tcolorbox}
\end{tcolorbox}
\caption{A complete evaluation from the semantic training cache, with the syntax-aware frontiers used for training. All 219 explicit tokens and all 28 latent frontiers are shown inside one frame. Contiguous background regions follow the coarse semantic hierarchy, their identities are written at the right, and each uniform dark \texttt{|} marks one latent frontier. }
\label{fig:complete-three-region-chunking}
\end{figure}

\begin{figure}[htbp]
\centering
\begin{tcolorbox}[
    enhanced,
    colback=fixedred!5!white,
    colframe=fixedred!70!black,
    boxrule=0.7pt,
    arc=1mm,
    boxsep=0pt,
    left=1.2mm,
    right=1.2mm,
    top=1mm,
    bottom=1mm
]
\begin{minipage}[c]{0.82\linewidth}
{\ttfamily\scriptsize\raggedright
--- Compliance Check ---\\
Identified Gatekeeper\fixedexamplebar{} Criterion: Criterion 1: The response\fixedexamplebar{} directly addresses the question posed by providing a\fixedexamplebar{} clear and explicit answer.\\
--- Analysis ---\fixedexamplebar{}\\[0.25em]
**Response A:**\\
- Criterion \fixedexamplebar{}1 [Hard Rule]: Met. The\fixedexamplebar{} response provides the explicit answer "7".\fixedexamplebar{}\\
- Criterion 2 [Principle]:\fixedexamplebar{} Met. The answer "7" is\fixedexamplebar{} self-contained and clear.\\
- Criterion \fixedexamplebar{}3 [Principle]: Met. The\fixedexamplebar{} response uses conventional notation (a numeral)\fixedexamplebar{} to communicate the answer.\\[0.25em]
**Response B\fixedexamplebar{}:**\\
- Criterion 1 [Hard\fixedexamplebar{} Rule]: Met. The response provides the\fixedexamplebar{} explicit answer "7".\\
- Criterion \fixedexamplebar{}2 [Principle]: Met. The\fixedexamplebar{} answer is presented in a complete sentence,\fixedexamplebar{} making it self-contained and clear.\\
-\fixedexamplebar{} Criterion 3 [Principle]: Met\fixedexamplebar{}. The response uses appropriate and conventional language\fixedexamplebar{} to communicate the answer effectively.\\[0.25em]
--- Final\fixedexamplebar{} Judgment ---\\
Justification: Both responses successfully\fixedexamplebar{} meet all criteria. Response B is slightly\fixedexamplebar{} better as it provides a more complete and\fixedexamplebar{} conversational answer, directly addressing the question\fixedexamplebar{} in a full sentence while still being concise\fixedexamplebar{} and clear.\fixedexamplebar{}\\[0.35em]
\textbf{Preference label (outside the chunked trace): Response B}
}\end{minipage}\hfill
\begin{minipage}[c]{0.14\linewidth}
\raggedleft\color{fixedred!60!black}\bfseries\footnotesize
Fixed-Rate\\[-0.1em]Chunking ($\rho=8$)
\end{minipage}
\end{tcolorbox}
\caption{Fixed-Rate Chunking of exactly the same evaluation trace as Figure~\ref{fig:complete-three-region-chunking}. The 219 explicit tokens are divided into 27 complete groups of eight and one final group of three, yielding the same 28 latent tokens. Only the red frontier locations change: unlike Semantic Chunking, they may split a phrase, a rubric criterion, or a transition between coarse semantic regions.}
\label{fig:complete-fixed-rate-chunking}
\end{figure}

%% file: sections/interpreter_appendix.tex
\section{Latent Trace Interpreter Details}
\label{app:lti-details}

The Latent Trace Interpreter (LTI) reconstructs an explicit evaluation from a latent trajectory. It is initialized from the LatentGRM-8B Stage~1 decoder, whose pretrained weights remain frozen, and learns its own LoRA adapter.

\subsection{Vocabulary-Probability Readout}

For each position $i$, the readout consumes top-$K$ vocabulary indices $S_i^{\mathrm{save}}$ and normalized weights $\bar q_i^{\mathrm{save}}$, following the notation in Section~\ref{sec:latent-interpreter}. For a teacher-derived trajectory, $(S_i^{\mathrm{save}},\bar q_i^{\mathrm{save}})=(S_i,\bar\alpha_i)$ is exported by the Stage~1 encoder. For an autonomous trajectory, $(S_i^{\mathrm{save}},\bar q_i^{\mathrm{save}})=(S_i^\theta,\widetilde q_i(\mathbf g))$ is saved from the inference recurrence in Equations~\ref{eq:appendix-online-topk}--\ref{eq:appendix-online-feedback}.

Equation~\ref{eq:interpreter-input-mixture} maps the saved vocabulary probabilities through the interpreter's embedding table. The input is $[x,\texttt{<think>},z^{\mathrm{int}}_{1:N}]$, where $x=(q,\mathcal R,a,b)$. Each mixture occupies one position in the original trajectory order, with consecutive position indices and causal attention. Text generation begins after $z_N^{\mathrm{int}}$.

\subsection{Full-Trajectory Reconstruction Objective}

We pair each cached trajectory with the complete teacher evaluation that produced it. Let $\omega$ denote the trainable LTI parameters, let $t=[c_{1:M},\texttt{</think>}]$, and let $L$ be its token length. The training sequence is
\begin{equation}
    [x,\texttt{<think>},z^{\mathrm{int}}_{1:N},c_{1:M},\texttt{</think>}],
    \label{eq:interpreter-training-sequence}
\end{equation}
with the token-averaged teacher-forcing objective
\begin{equation}
    \mathcal L_{\mathrm{int}}(\omega)=
    -\frac{1}{L}\sum_{j=1}^{L}
    \log p_\omega(t_j\mid x,\texttt{<think>},z^{\mathrm{int}}_{1:N},t_{<j}).
    \label{eq:interpreter-full-objective}
\end{equation}
Loss covers the complete evaluation and closing marker; prompt and latent positions are masked, and the binary preference label is excluded. Each example supplies one full trajectory. LTI thus reconstructs the evaluation from its beginning, whereas the Stage~1 decoder predicts the suffix after a sampled frontier.

\subsection{LTI Training Settings}

For the reconstruction study, the LTI adapter is optimized with Equation~\ref{eq:interpreter-full-objective}; the frozen weights include the input embeddings. Table~\ref{tab:interpreter-settings} summarizes its training configuration.

\begin{table}[htbp]
    \centering
    \small
    \caption{LTI training configuration. Effective batch size includes data-parallel workers and gradient accumulation.}
    \label{tab:interpreter-settings}
    \begin{tabular}{p{0.31\linewidth}p{0.61\linewidth}}
        \toprule
        \textbf{Setting} & \textbf{Value} \\
        \midrule
        Initialization & LatentGRM-8B Stage~1 decoder; frozen backbone \\
        Training data & 28,430 examples; 512 validation examples; seed 42 \\
        Trajectory view & Full trajectory in forward order; one view per example \\
        LTI representation & Probability-weighted top-$K$ input embeddings \\
        LoRA & Rank 32; $\alpha=64$; dropout 0.05; all attention and MLP projections \\
        Training duration & 2 epochs \\
        Learning rate / effective batch & $10^{-5}$ / 32 \\
        Per-device batch / accumulation & 1 example / 4 steps \\
        Optimizer and schedule & AdamW; cosine decay; 3\% warmup; zero weight decay \\
        Precision / sequence length & bfloat16; gradient checkpointing; maximum 6,144 tokens \\
        Hardware & $8\times$ NVIDIA H20 \\
        \bottomrule
    \end{tabular}
\end{table}

\FloatBarrier

%% file: sections/latent_mechanism_appendix.tex
\subsection{Protocol for Rubric-Based Process Evaluation}
\label{app:mechanism-protocol}

This appendix specifies the LatentGRM-8B interventions in Section~\ref{sec:rubric-analysis} and supplementary trajectory analyses.

\subsubsection{Data and Annotation}

\begin{table}[H]
\centering
\small
\renewcommand{\arraystretch}{1.15}
\begin{tabular}{@{}p{0.17\linewidth}p{0.79\linewidth}@{}}
\toprule
\textbf{Component} & \textbf{Text from the accepted example} \\
\midrule
Request & Who created the Superman cartoon character? \\
Response A & Superman, the iconic comic book superhero, was created by writer Jerry Siegel and artist Joe Shuster. Superman first appeared in Action Comics \#1, which was published by Detective Comics, Inc. (later DC Comics) in June 1938. The character's immense popularity established him as one of the most enduring and recognizable figures in the superhero genre. \\
Response B & The Superman cartoon character was created by Jerry Siegel and Joe Shuster in 1938. \\
\midrule
Original & 3. The response should be concise and avoid unnecessary elaboration. [Principle] \\
Paraphrase & 3. The response should be brief and avoid including extraneous information. [Principle] \\
Counterfactual & 3. The response should provide thorough and enriching details that go beyond the minimum facts required to answer. [Principle] \\
\bottomrule
\end{tabular}
\caption{Example 1: brevity versus contextual detail.}
\label{tab:rq1-criterion-example-2}
\end{table}

\begin{table}[H]
\centering
\small
\renewcommand{\arraystretch}{1.15}
\begin{tabular}{@{}p{0.17\linewidth}p{0.79\linewidth}@{}}
\toprule
\textbf{Component} & \textbf{Text from the accepted example} \\
\midrule
Request & Generate an advertisement slogan that promotes healthy eating. \\
Response A & ``Get Fit, Fuel Up - Find Balance with Our App!'' \\
Response B & Eat healthy, feel healthy! \\
\midrule
Original & 5. The response should employ vivid, engaging imagery or metaphors to capture attention. [Principle] \\
Paraphrase & 5. The response should use striking, imaginative language or figurative expressions to draw attention. [Principle] \\
Counterfactual & 5. The response should use simple, literal language that directly communicates its message without figurative embellishment. [Principle] \\
\bottomrule
\end{tabular}
\caption{Example 2: figurative versus literal language.}
\label{tab:rq1-criterion-example-3}
\end{table}

\paragraph{Sampling.}
Using seed 42, we sampled 200 distinct question and response pairs from each of RewardBench Chat and Chat Hard, balancing the reference winner's displayed position within each domain. Semantic validation produced 398 accepted pairs, comprising 198 Chat and 200 Chat Hard pairs. Each pair is evaluated in both response orders, and intervention scores are mapped to response identity before aggregation.

\paragraph{Criterion generation and verification.}
GLM-5.2 was called separately for two distinct roles: generation and verification. In the generation call, it proposed an equivalent paraphrase and a counterfactual for one criterion, while all other text remained fixed. In the verification call, it independently checked the proposed edits against the rubric without benchmark labels or judge outputs. We retain for which the beneficiary inferred in the generation stage agrees with the beneficiary inferred in the verification stage.

\paragraph{Examples of GLM-5.2 criterion edits.}
Tables~\ref{tab:rq1-criterion-example-2} and \ref{tab:rq1-criterion-example-3} show two accepted edits from the 398-pair analysis. Requests, responses, and criterion texts are reproduced verbatim; only the indicated criterion changes. Directions below are the proposer and reviewer's agreed criterion-level predictions, not measured judge outcomes or overall winner labels.

\subsubsection{Judge Analysis}

For each pair, we build three rubric conditions: the original rubric $\mathcal{R}_0$, a paraphrase control $\mathcal{R}_p$ that rewords one criterion without changing its meaning, and a criterion intervention $\mathcal{R}_c$ that changes that criterion to favor a designated response. The paraphrase condition serves as a control: if a semantics-preserving rewording also shifted the preference, the effect could not be attributed to the criterion's meaning. We index the trajectory source by $s\in\{0,p,c\}$, meaning the trajectory was generated under $\mathcal{R}_s$.

For each source $s\in\{0,p,c\}$, we probe its latent trajectory at $0\%,10\%,\ldots,100\%$ of its recorded length. This matches relative progress across sources without forcing equal lengths. At each probe, we append the latent end marker and a shared \texttt{Response} prefix, and score the remaining label tokens. We measure the log-odds difference $D_i(C)$ between the two response labels (Equation~\ref{eq:preference-logprob-difference}).

\paragraph{Criterion intervention.}
Each rubric generates its own autonomous trajectory. For $s\in\{p,c\}$, we compare
\begin{equation}
 \Delta D_{i,s}(j)=
 D_i(\mathcal R_s,Z_s^{\leq k_{i,s,j}})-
 D_i(\mathcal R_0,Z_0^{\leq k_{i,0,j}}).
\end{equation}
This contrast combines prompt and trajectory effects; its zero-latent value captures a difference already present in the rubric prompt.

\paragraph{Source replacement.}
We fix the recipient rubric to $\mathcal R_0$ and replay the source trajectory generated under $\mathcal R_s$ in a fresh baseline context. Donor prompt tokens and donor KV caches are not transferred. For $s\in\{p,c\}$, we compare
\begin{equation}
 \Delta D_{i,s}(j)=
 D_i(\mathcal R_0,Z_s^{\leq k_{i,s,j}})-
 D_i(\mathcal R_0,Z_0^{\leq k_{i,0,j}}).
\end{equation}

We compute signed effects per pair/order, average the two orders within each pair, then average pairs within each domain and weight the domains equally. The interventions establish criterion-directed preference shifts and their transfer through latent prefixes.

%% file: sections/direct_judge_appendix.tex
\subsection{Direct Judgment and Voting}
\label{app:direct-judge}

We compare DirectJudge, explicit Rubric-RM, and LatentGRM using Qwen3-8B on the RewardBench Chat and Chat Hard domains and the RewardBench2 Precise IF and Focus domains. DirectJudge is initialized from the same backbone and trained for two epochs on the same OpenRubrics Dataset, supervising only the final \texttt{Response A/B} verdict. It receives the request, rubric, and both responses, but generates no intermediate evaluation trace.

\begin{table}[ht]
    \centering
    \small
    \setlength{\tabcolsep}{6pt}
    \caption{\textbf{Direct judgment versus explicit and latent evaluation (\%).}}
    \label{tab:direct-judge-voting}
    \begin{tabular}{llrrrr}
        \toprule
        & & \multicolumn{2}{c}{\textbf{RewardBench}} & \multicolumn{2}{c}{\textbf{RewardBench2}} \\
        \cmidrule(lr){3-4}\cmidrule(lr){5-6}
        \textbf{Model} & \textbf{Votes} & \textbf{Chat} & \textbf{Chat Hard} & \textbf{Precise IF} & \textbf{Focus} \\
        \midrule
        DirectJudge & 1 & 90.92 & 67.21 & 35.94 & 85.15 \\
                    & 5 & 91.62 & 67.43 & 41.56 & 86.57 \\
                    & 9 & 90.78 & 67.56 & 41.25 & 86.47 \\
        \midrule
        Rubric-RM   & 1 & 89.80 & 69.80 & 31.90 & 77.90 \\
                    & 5 & \best{91.80} & 70.50 & 45.00 & 81.10 \\
                    & 9 & 91.78 & 70.39 & \best{45.68} & 82.53 \\
        \midrule
        LatentGRM   & 1 & 88.70 & 70.50 & 37.90 & 84.60 \\
                    & 5 & 89.90 & 73.00 & 43.40 & 87.50 \\
                    & 9 & 90.08 & \best{73.24} & 43.50 & \best{87.68} \\
        \bottomrule
    \end{tabular}
\end{table}

Direct judgment is a strong low-decoding-cost baseline. Table~\ref{tab:direct-judge-voting} shows competitive Chat performance and strong Focus scores, including an advantage over Rubric-RM. On Chat Hard, however, DirectJudge reaches only 67.43\% at vote@5, compared with 70.50\% for Rubric-RM and 73.00\% for LatentGRM, and increasing votes from one to five improves it by only 0.22 points. This gap shows that direct judgment is limited on challenging preference comparisons.

DirectJudge produces a terminal assessment without an autoregressive evaluation trajectory, and thus loses the test-time scaling enabled by sampling multiple evaluation paths. It also lacks the diversity needed for reinforcement learning to optimize. LatentGRM instead retains a compact continuous trajectory that can be probed, inspected, and potentially optimized, at lower generation cost than explicit evaluation.

%% file: sections/revision_appendix.tex
\subsection{Data Revision Protocol and Evaluation Details}
\label{app:revision-utility}

An evaluation model uses the full judging context, including the task-specific rubric, to produce feedback $F$ for a candidate response $x$. A separate reviser then generates
\begin{equation}
 G(q,x,F)\rightarrow x'.
\end{equation}
The reviser receives only the question, the response to revise, and the feedback interface under study; the rubric is never included in its input.

We evaluate 240 RewardBench Rubric examples, with 120 from Chat and 120 from Chat Hard.  The rubric is used to generate evaluation feedback and to score the resulting revision, but is not exposed to the reviser.  All four conditions use the same Qwen3-8B Base reviser in a fresh context:
\begin{enumerate}
    \item \textbf{Prompt-only}: the reviser receives only the question and the response to revise.
    \item \textbf{Explicit trajectory}: the textual reasoning generated by the Rubric-RM-8B.
    \item \textbf{Latent only}: the complete LatentGRM trajectory, represented at every recurrent step by the probability-weighted mixture of its top-K vocabulary embeddings, is inserted directly in the reviser's context. This is a training-free interface: the reviser has never been trained to decode or consume LatentGRM trajectories. It measures whether a pretrained model can directly turn latent reasoning into a better response.
    \item \textbf{Interpreter-restored trajectory}: LTI converts the same soft latent trajectory into textual feedback, which is then supplied to a fresh frozen reviser context.
\end{enumerate}

We use DeepSeek-V4.1-Flash~\citep{deepseekai2026deepseekv41flash} as the external evaluation model. For each original or revised answer, it sees the question, the complete rubric. It independently marks every rubric criterion as satisfied or unsatisfied and provides brief supporting evidence. The original 240 answers contain 1,973 criterion decisions: 1,076 satisfied and 897 unsatisfied. On a preselected 20-example subset, reversing criterion order and paraphrasing the judging instruction yields 93.76\% criterion-level agreement and Cohen's $\kappa=0.803$.

For original labels $s_{ij}$ and revised labels $s'_{ij}$, we measure
\begin{align}
    \operatorname{Repair} &= \frac{\sum_{ij}\mathbf{1}[\neg s_{ij}\wedge s'_{ij}]}{\sum_{ij}\mathbf{1}[\neg s_{ij}]},\\
    \operatorname{Regression} &= \frac{\sum_{ij}\mathbf{1}[s_{ij}\wedge \neg s'_{ij}]}{\sum_{ij}\mathbf{1}[s_{ij}]},
\end{align}
the percentage of answers satisfying every rubric criterion (Full), and overall criterion satisfaction.